\documentclass[10pt,journal,compsoc]{IEEEtran}

\ifCLASSOPTIONcompsoc
  \usepackage[nocompress]{cite}
\else
  \usepackage{cite}
\fi

\ifCLASSINFOpdf
  
\else
  
\fi

\newcommand{\mon}[1]{%
   \expandafter\monAUX\expandafter{\detokenize{#1}}%
}

\newcommand*\patchAmsMathEnvironmentForLineno[1]{%
\expandafter\let\csname old#1\expandafter\endcsname\csname #1\endcsname
\expandafter\let\csname oldend#1\expandafter\endcsname\csname end#1\endcsname
\renewenvironment{#1}%
{\linenomath\csname old#1\endcsname}%
{\csname oldend#1\endcsname\endlinenomath}}%
\newcommand*\patchBothAmsMathEnvironmentsForLineno[1]{%
\patchAmsMathEnvironmentForLineno{#1}%
\patchAmsMathEnvironmentForLineno{#1*}}%
\AtBeginDocument{%
\patchBothAmsMathEnvironmentsForLineno{equation}%
\patchBothAmsMathEnvironmentsForLineno{align}%
\patchBothAmsMathEnvironmentsForLineno{flalign}%
\patchBothAmsMathEnvironmentsForLineno{alignat}%
\patchBothAmsMathEnvironmentsForLineno{gather}%
\patchBothAmsMathEnvironmentsForLineno{multline}%
}

\usepackage{amsmath}
\usepackage{graphicx}
\usepackage{lineno}
\usepackage{array}
\usepackage{longtable}
\usepackage[usenames,dvipsnames]{color}
\usepackage{amssymb}
\usepackage{textcomp}
\usepackage[pagebackref=true,breaklinks=true,colorlinks,citecolor=green]{hyperref}
\usepackage{url}
\usepackage{float}
\usepackage{multirow}
\usepackage{microtype}
\usepackage{bm}

\usepackage{times}
\usepackage{epsfig}
\usepackage{graphicx}
\usepackage{amsmath}
\usepackage{amssymb}

\usepackage{subfig}
\usepackage{gensymb} 
\usepackage{booktabs}
\usepackage{multirow}
\usepackage{enumitem}
\usepackage{caption}
\usepackage{nccmath}
\usepackage{multirow, makecell}

\def\rot90{\rotatebox{90}}

\definecolor{col_bicycle}{RGB}{245, 230, 100}
\definecolor{col_person}{RGB}{255,   125, 125}
\definecolor{col_sidewalk}{RGB}{75,0,75}
\definecolor{col_parking}{RGB}{255,150,255}
\definecolor{col_building}{RGB}{0,200,255}
\definecolor{col_other_veh}{RGB}{255,   190,   90}

\begin{document}

\title{Sequential Point Clouds: A Survey}

\author{Haiyan Wang
        and~Yingli Tian,~\IEEEmembership{Fellow,~IEEE,}
\IEEEcompsocitemizethanks{
\IEEEcompsocthanksitem Haiyan Wang is with the Department
of Electrical Engineering, The City College of New York, New York,
NY, 10031.\protect\\
E-mail: hwang3@ccny.cuny.edu
\IEEEcompsocthanksitem Yingli Tian (Corresponding author) is with the Department of Electrical Engineering, The City College, and the Department of Computer Science, the Graduate Center, the City University of New York, New York, NY, 10031.\protect \\
E-mail: ytian@ccny.cuny.edu
}
\thanks{This material is based upon work supported by the National Science Foundation under award number IIS-2041307.}}

\markboth{Journal of \LaTeX\ Class Files}%
{Shell \MakeLowercase{\textit{et al.}}: Bare Advanced Demo of IEEEtran.cls for IEEE Computer Society Journals}

\IEEEtitleabstractindextext{%
\begin{abstract}
Point cloud has drawn more and more research attention as well as real-world applications. However, many of these applications (e.g. autonomous driving and robotic manipulation) are actually based on sequential point clouds (i.e. four dimensions) because the information of the static point cloud data could provide is still limited. Recently, researchers put more and more effort into sequential point clouds. This paper presents an extensive review of the deep learning-based  methods for sequential point cloud research including dynamic flow estimation, object detection \& tracking, point cloud segmentation, and point cloud forecasting. This paper further summarizes and compares the quantitative results of the reviewed methods over the public benchmark datasets. Finally, this paper is concluded by discussing the challenges in the current sequential point cloud research and pointing out insightful potential future research directions.  

\end{abstract}

\begin{IEEEkeywords}
Point cloud; Sequential data; Deep learning; Flow estimation; Object detection \& tracking; Point cloud segmentation; Point cloud forecasting.
\end{IEEEkeywords}}

\maketitle

\IEEEdisplaynontitleabstractindextext

\IEEEpeerreviewmaketitle

\ifCLASSOPTIONcompsoc
\IEEEraisesectionheading{\section{Introduction }\label{sec:introduction}}
\else

\fi

\IEEEPARstart{W}{ith} the development of recent deep learning and sensor technologies, the expense of 3D point cloud acquisition has significantly dropped. Point cloud data can be easily captured through 3D scanners, Lidars, or RGBD cameras, which comprise a set of unordered points represented by  \textit{XYZ} in world coordinates with permutation invariant properties. Compared to other data formats, like 2D image, even 3D voxel or mesh, the point cloud is a more practical data representation for our real world. 2D images lose the spatial geometric information of 3D space, while other grid-based representations (e.g. voxel and mesh) suffer from the redundancy of the inner space representation and massive computation.

Recent research efforts have made great contributions to the static point cloud learning process. The survey paper \cite{Guo2019DeepLF} provided an elaborate summary of 3D point cloud learning methods including various downstream tasks and applications. Basically, some methods just pursued an easier deep learning way which employs the convolution operation on the high dimension 3D data \cite{VoxelNet,point-voxel,voxsegnet}. These methods usually require transferring point cloud to other regular data formats such as voxel or mesh representations. The input of grid data representation makes it possible to extend the idea of advanced 2D convolution network design to the 3D domain for  high-level feature extraction. Although the convolution is attractive, these methods suffer a lot from heavy computation costs and quantization errors due to the grid representation. 
The seminal work PointNet \cite{PointNet} and PointNet++ \cite{PointNet++} introduced a straightforward solution based on raw point cloud input and extracted high-level feature representations through  novel sampling and grouping strategies. Inspired by these two pioneer methods, tremendous studies developed more and more advanced structures and achieved impressive performance on different computer vision applications. These direct point-based methods usually maximally preserved the 3D geometry information of the input data  and well balanced the efficiency and efficacy. 

\begin{figure}[t!]
\centering
\includegraphics[width=0.48\textwidth]{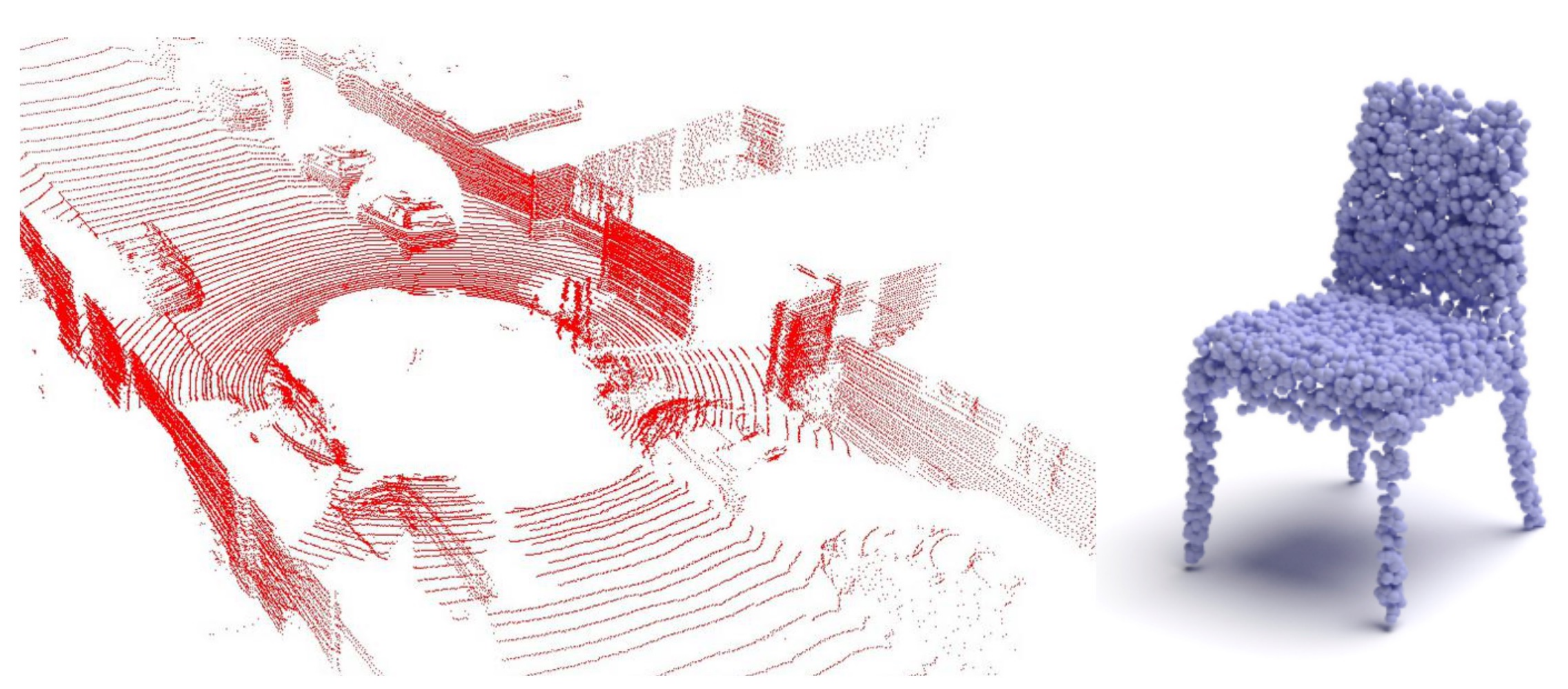}
\caption{Demonstration of sequential point cloud (left) and static point cloud (right). The figure is from \cite{weng2020unsupervised} with author’s permission. }
\label{fig:motivation_1}
\end{figure}

\begin{figure*}[ht!]
\centering
\includegraphics[width=1\textwidth]{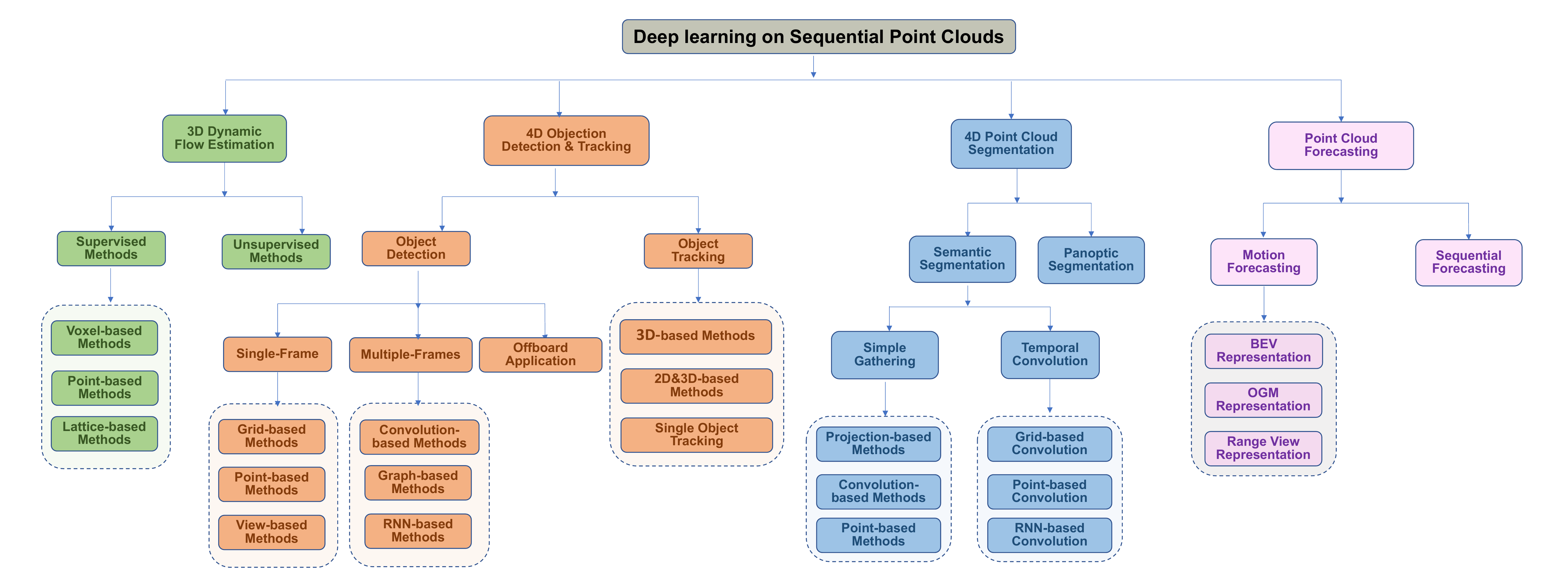}
\caption{A taxonomy of deep learning methods for sequential point cloud.}
\label{fig:taxonomy}
\end{figure*}

However, even with remarkable representations of the static point cloud, it is insufficient to fully represent our real world especially when there are motions. The dynamic real world is actually with three spatial dimensions plus one time dimension (i.e. 4D), which leads to a huge uncertainty compared with the single static point cloud. The features of the scene or objects may change along the time sequence causing the potential missing, occlusion, or unseen information. Even these uncertainties are inevitable in our dynamic world, they are critical to be aware of and estimated especially in real-world applications such as self-driving or AR/VR techniques. Thus, many deep learning tasks (e.g. dynamic flow estimation, object detection \& tracking, point cloud segmentation, and point cloud forecasting, etc.) are worth to explore for learning the spatio-temporal information from 4D sequential point cloud (SPL) data. In a short period, the motion information such as point flow which is similar to 2D optical flow can be estimated based on consecutive point cloud frames. Also, based on the previous several frames, the point cloud of the future moment can be predicted which is applied by vast kinds of forecasting tasks such motion and sequential forecasting. The point cloud generation falls in this application as well. The recent thriving tasks such as object tracking, action recognition, 4D point cloud reconstruction, and even the 4D segmentation can also benefit from the long-time temporal information embedded in the point cloud sequence. Motivated by the distinguished property of the sequential point cloud and these popular applications, the research focuses are diverting from the static point cloud to the dynamic sequential point cloud.

Sequential point cloud (SPL) is defined as a sequence of static point cloud frames $S = S_1, S_2, ...S_t, ... , S_T, (t = 1,2,...,T )$ where $T$ is the time length. Each point cloud frame $S_t$ consists of a set of unordered points which are permutation invariant $S_t = {p_1, p_2,...,p_n,..., p_N} $, $N$ is the number of points for the point cloud frame $S_t$. The point $p_n$ inside $S_t$ is represented with both 3D location $X_n \in \mathbb{R}^3$ and feature vector $F_n \in \mathbb{R}^c$. As shown in Figure \ref{fig:motivation_1}, the left one demonstrates a SPL of an outdoor scene and the right one indicates a static point cloud of a single object. Compared to static point cloud, SPL is unique with the following properties:

\begin{itemize}
    \item \textbf{Large scale}. A static scene point cloud normally contains plenty of points and can easily reach a scale of millions. SPL unites a sequence of static point clouds, the number of points are extremely immense. 

    \item \textbf{Permutation invariant of single frame}. Every single scan in SPL is a set of unordered points which is invariant to any permutation and geometric transformation such as translation or rotation. These operations will not alter the point cloud properties or classification results.  
    
    \item \textbf{Permutation variant for multiple frames}. Among multiple frames of point clouds, the order of these frames is the most critical characteristic which makes it distinctive. It reflects the temporal information along with the time series including the dynamic motion and deformation of the object in the point clouds. 

    \item \textbf{4D Contextual Correlation / Continuum}. The learning of SPL ought not to separate the spatial and the temporal. Instead, for the 4D continuum, a spatio-temporal correlation structure contains extremely rich contextual information availing a better scene understanding compared to the single static point cloud. 
\end{itemize}

Despite the superior properties and importance of SPL, it is especially challenging to process 4D data in an effective and efficient manner due to the large scale and sophistication of the spatio-temporal relations between multiple frames.
The core idea of processing 4D sequential point cloud data is to take benefit of both spatial and temporal dimensions. Meanwhile, the way of extracting and merging temporal information is essential during this process. 
Many methods have been developed and achieved impressive performance on 3D static point clouds. 

There are some reviews before that recapped the deep learning methods on the general 3D data \cite{3D:Deep:Learning:Survey,ahmed2018deep,Xie2019review,rahman2019recent} or especially to the static point cloud \cite{Guo2019DeepLF,Liu2019DeepLO}. However, none of them focus on modeling  SPL. This paper presents an extensive review of the deep learning-based methods for 4D SPL research and emphasizes the temporal encoding and modeling of the spatio-temporal correlation structure. As shown in Figure \ref{fig:taxonomy}, we provide a thorough comparison of the existing methods on the public benchmark datasets for diverse tasks and applications including dynamic flow estimation, object detection $\&$ tracking, point cloud segmentation, and point cloud forecasting.  We further summarize the research challenges of the SPL and point out several trends for future research.

The rest of the paper is organized as follows. Sec. \ref{sec:network} introduces the common deep network architectures employed to process SPL. Sec. \ref{sec:datasets} lists the benchmark datasets for various applications of the SPL data. The downstream tasks of different applications of SPL are summarized in Sec. \ref{sec:flow} for scene flow estimation, Sec. \ref{sec:object_detection} for objection detection, Sec. \ref{sec:object_tracking} for object detection and tracking, Sec. \ref{sec:object_segmentation} for object segmentation, and Sec. \ref{sec:forecasting} for point cloud forecasting. Sec. \ref{sec:future} provides a few potential future research directions on SPL and Sec. \ref{sec:conclusion} concludes the whole survey.

\section{Common Deep Network Architectures  } \label{sec:network}

Before diving into the specific deep learning tasks, in this section, we briefly summarize the common deep networks for general feature learning of high dimensional data. There are mainly two streams methods. One is applying the convolution neural networks to directly learn the spatio-temporal features (Sec. \ref{sec:2:convolution}). Another stream is adopting the recurrent networks and recurrently comprehending the hidden states (Sec. \ref{sec:2:rnn}). 

\subsection{Convolution Neural Network} \label{sec:2:convolution}

According to the representation of the input data, these methods can be categorized into grid-based and point-based architectures.

\subsubsection{Grid-based Architectures}
\label{sec:net:grid}

\begin{figure}[H]
\centering
\includegraphics[width=0.48\textwidth]{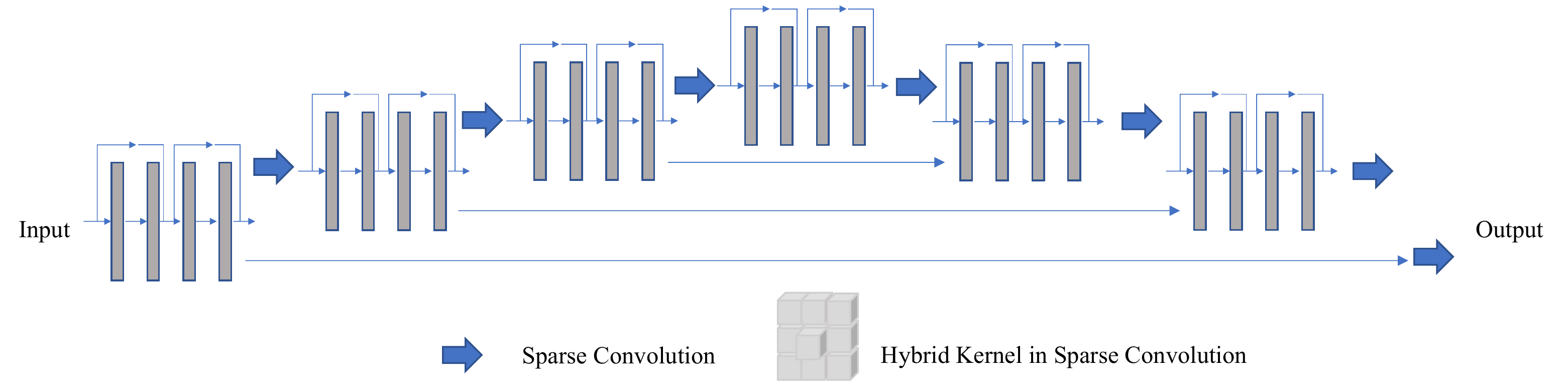}
\caption{The illustration of a Grid-based Architecture. Figure is reproduced based on \cite{MinkNet}.}
\label{fig:2:grid}
\end{figure}

These methods transfer point clouds into regular representations such as voxel or point tube, which could further support the common convolution layers to extract features. Figure \ref{fig:2:grid} shows a standard grid-based network for feature learning.

\noindent\textbf{4D MinkNet} \cite{MinkNet} was the first one to exploit the common deep learning network on high-dimensional data. It adopted the idea from Sparse Tensor \cite{graham2014spatially} and proposed a generalized sparse convolution to operate SPL. The proposed convolution layer can be blended with various deep networks and well generalized to different tasks. To deal with the computational problem when generalizing the convolution to high dimensional spaces, the authors designed a novel kernel that is not hyper cubic and thus diminishes the memory cost. Moreover, the high-dimensional conditional random fields were introduced to enforce the consistency between the space and time domains. Incorporating all of these designs, MinkNet was established to provide a general deep network to handle sequential point clouds just as the 2D perception. 

\noindent\textbf{PSTNet} \cite{fan2021pstnet} was another grid-based method which performs spatio-temporal convolution on the sequential point clouds. They decoupled the spatial and temporal information from the input raw point clouds which is shown to be more reasonable and effective. Unlike the above-mentioned method MinkNet \cite{MinkNet}, which suffered from a large amount of computation due to the voxel representation, PSTNet developed a Point tube structure to manage the input data and conduct the proposed convolution. The point tube incorporated the spatial and temporal kernels separately to capture the spatio-temporal local structure information. To tackle both the sequence-level and point-level classification tasks such as semantic segmentation, the authors introduced the PST convolutions and transposed convolutions to construct the PSTNet hierarchically.

\subsubsection{Point-based Architectures}
\label{sec:net:point}

\begin{figure}[H]
\centering
\includegraphics[width=0.48\textwidth]{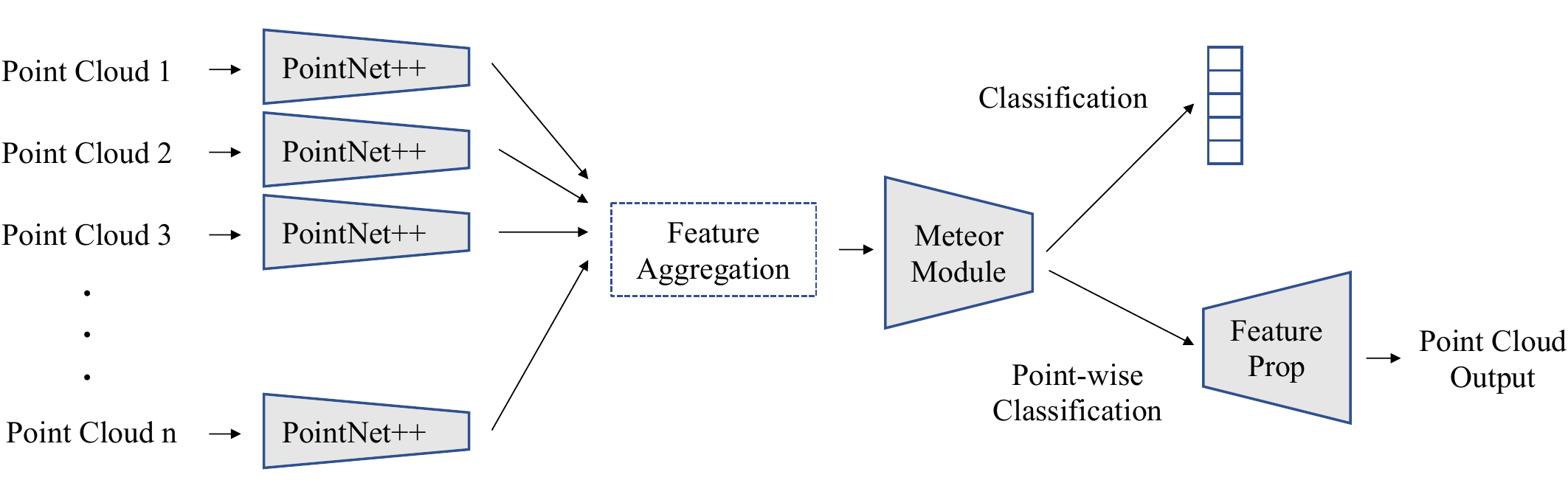}
\caption{The illustration of a point-based architecture build upon \cite{MeteorNet}.}
\label{fig:2:point}
\end{figure}

These methods assemble the network based on a set of MLP (Multiple Layer Perception) layers and aggregate feature information from neighborhood regions along with both spatial and temporal domains. We show a point-based architecture in Figure \ref{fig:2:point}.

\noindent\textbf{MeteorNet} \cite{MeteorNet} was a seminal work that explores deep learning on the SPL data with a direct point-based method. Treating PointNet++ \cite{PointNet++} as an elementary unit, it proposed an architecture to digest the input point cloud sequences by early/later fusion methods, which provided a common solution to learn sequential point cloud features. Moreover, another core contribution was that they fused the temporal information by explicitly grouping the meaningful neighbor regions. Two grouping methods were proposed to solve the problem, direct grouping and chain-flow grouping. The direct grouping-based method increased the radius monotonically with the time increases to search the nearest neighbor region. The chain-flow-based method predicted the flow using the offline network Flownet3D \cite{liu2019flownet3d} to better find the correspondence between frames. The scene flow could help to find a better search radius to confirm the nearest neighbor region.

\subsection{Recurrent Neural Network} \label{sec:2:rnn}

Besides the convolution networks, the recurrent neural network is another intuitive method to process SPL data. PointRNN \cite{fan2019pointrnn} proposed a Point Recurrent Neural Network (PointRNN) to learn the representation of moving point clouds. PointRNN extended the idea from the 2D RNN to the 3D/4D RNN. Meanwhile, due to the problem of representing the point cloud into a single state vector, PointRNN separated the position features and auxiliary features to sever as the state vectors for updating. Another problem when applying RNN to  point clouds is that simple concatenation of data over the temporal series cannot be conducted due to the point clouds being unordered. Instead, PointRNN tackled this problem by adopting a correlation layer between the previous state vector and current input data. It would search the nearest neighbor points linked to the query point and concatenate them separately. The final pooling operation would extract a single vector from the previous representations. The authors affirmed the effectiveness of LSTM for the moving point cloud prediction task.

\subsection{Discussion}

All of the above-mentioned deep learning methods investigated  general pipelines to conduct feature learning from SPL data. Here we briefly summarize the characters of different network architectures:  

\begin{itemize}
    \item Convolution Neural Networks exploit operations over the entire spatial and temporal domain. The extracted spatial and temporal features have a more mutual impact. Thus these networks focus more on the feature consistency along temporal sequence. Some high-level tasks which require a better semantic understanding such as detection and segmentation will be benefited more from feature learning of Convolution Neural Network. 
    
    \item Essentially, Recurrent Neural Networks emphasize more on the long-range dependency along the time dimension. The temporal relation between distinct frames is explicitly represented by their recurrent design. Thus, those long-range sequence tasks are more appropriate with Recurrent Neural Network such as action recognition or object tracking.

\end{itemize}

\section{Datasets} \label{sec:datasets}

We summarize the datasets commonly used in SPL analysis in Table \ref{tab:datasets}.
\begin{table*}[ht]
\centering
\caption{Summary of the commonly used SPL datasets.}
\label{tab:datasets}
\resizebox{1\textwidth}{!}{
\begin{tabular}{c||c|c|c|c}
\hline
Datasets           & Size                       & Annotation                    & Train$\&$Val/Test & Synthetic  \\ \hline\hline
ATG4D    \cite{meyer2019lasernet}          & 5,500  sequences, 1.2M frames & Motion trajectory                  & 5,000/500 (Frame)     & $\times$  \\
Flythings3D \cite{Mayer2016ALD}       & 22,000  pair frames   & Point-wise scene flow         & 20,000/2,000 (Frame)   & \checkmark \\
KITTI     \cite{geiger2012we}         & 21  sequences, 7,987 frames   & Bounding box; Object ID         & 16/5  (Sequence)        & $\times$   \\
KITTI Raw    \cite{geiger2013vision}       & 151  sequences  & Raw data (Point clouds)             &  106/45  (Sequence)            &    $\times$        \\
KITTI Scene Flow  \cite{KITTI:Scene:flow} & 150 pair frames     & Point-wise scene flow                 & 100/50 (Pair)       & $\times$   \\
NuScenes Dataset \cite{caesar2020nuscenes}   & 1,000 scenes, 0.3M frames    & Bounding box; Motion trajectory  & 850/150 (Scene)     & $\times$   \\
SemanticKitti  \cite{semantickitti}    & 22 sequences, 43,551  frames   & Point-wise class label; Object ID        & 19,130/4,071 (Frame)   & $\times$   \\
Synthia 4D  \cite{ros2016synthia}        & 6  sequences, 22,589 frames    & Point-wise class label          & 20,703/1,886  (Frame)       & \checkmark \\

Waymo Open Dataset \cite{sun2020scalability} & 1,150  sequences, 20M frames & Bounding box; Scene flow                    & 1,000/150 (Sequence)     & $\times$   \\ \hline

\end{tabular}}
\end{table*}

\noindent\textbf{ATG4D Dataset} \cite{meyer2019lasernet} consists of 5,000 sequences of training data including total 1.2 million Lidar sweeps, while the testing set contains 500 sequences and 5,969 sweeps. The dataset is captured by the Velodyne 64E LiDAR and is mainly used for motion forecasting tasks.

\noindent\textbf{Flythings3D Dataset} \cite{Mayer2016ALD}
 is currently the largest synthetic dataset for scene flow estimation which contains about 22,000 stereo frames with spatial size of 960x540 pixels. These images are rendered from randomized synthetic sequences along with moving objects from ShapeNet \cite{3d:shapenets}.  The ground truth annotations include segmentation maps, disparity maps, disparity changes, and optical flow maps. Thus, the point cloud sequences can be reconstructed from the disparity maps and the corresponding groundtruth scene flow annotations can be obtained through back-projecting the 2D optical flow maps and disparity changes to the 3D space.
 
\noindent\textbf{KITTI Dataset} \cite{geiger2012we} is popular dataset and widely used for object detection and tracking tasks. It consists a total of 50 sequences with the split of 16 training sequences, 5 offline testing sequences and 29 online testing sequences. Specifically, sequence 0 to 15 are adopted for training and sequence 16 to 20 for offline testing.   

\noindent\textbf{KITTI Raw Dataset} \cite{geiger2013vision} is adopted for the sequential foresting task which is a superset for other KITTI versions such KITTI Scene flow, detection, etc. The raw KITTI dataset contains a total of 151 sequences of Lidar data, which is divided into 60, 46, and 45 for train, val, and test respectively. 

\noindent \textbf{KITTI Scene Flow 2015 Dataset} \cite{KITTI:Scene:flow}
 is a real dataset that is proposed/designed for autonomous driving along with the deep learning tasks such as flow estimation, SLAM (Simultaneous Localization and Mapping), semantic \& instance segmentation, depth prediction \& completion, and object detection \& tracking. The dataset has been collected around the city of Karlsruhe, Germany using RGBD cameras and a Velodyne 64 LIDAR scanner. Based on KITTI dataset  \cite{geiger2012we}, Menze et al. \cite{KITTI:Scene:flow} took advantage of raw data and augmented it with detailed 3D CAD models, leading to a KITTI scene flow estimation benchmark with annotated groundtruth. There are a total of 200 training and 200 test scenes and previous research \cite{dewan2016rigid,ushani2017learning,liu2019flownet3d}  removed the useless points belong to ground to better focus on the scene flow information. Since there is no groundtruth annotation in the testing dataset, researchers often choose 150 out of 200 training scenes as their training set and the rest 50 as their testing set. 

\noindent\textbf{NuScenes Dataset} \cite{caesar2020nuscenes} is a large-scale autonomous driving dataset collected by Motional, with the purpose of aiding computer vision research and improving the safety of self-driving. The whole dataset consists of 1000 outdoor scenes where 850 scenes are used for training and validation, and the rest 150 scenes are for testing. There are a total of 1.4M object bounding boxes in 40k keyframes which are 7 times more object annotations than KITTI dataset \cite{geiger2012we}. The segmentation annotation covers 32 semantic categories resulting 1.4 billion annotated points across 40,000 pointclouds.

\noindent\textbf{SemanticKitti} \cite{semantickitti} is built upon KITTI \cite{geiger2012we} Odometry dataset, SemanticKitti is a large-scale dataset containing semantic annotations for sequential point clouds. The Lidar point clouds are scanned at a rate of 10 Hz which help to better understand both semantic and temporal information. The whole dataset consists of 22 point clouds sequences and 43,551 point cloud frames. Specifically, they are divided with 19,130 frames (sequence 0 to 10) for training, 4,071 frames (sequence 8) for validating and 20,351 frames (sequence 11 to 21) for testing. They provide challenges for both 3D and 4D semantic segmentation. There are a total of 25 object classes and the 3D semantic segmentation task only evaluates the performance of the 19 classes which are all static scenes or objects, while the 4D semantic segmentation task involves the 6 more moving classes leading to a more challenging situation. The temporal information between multiple frames is crucial to obtain better performance on the 4D semantic segmentation task.

\noindent\textbf{Synthia 4D Dataset} \cite{MeteorNet} Synthia \cite{ros2016synthia} is a large synthetic dataset collected for scene understanding, self-driving, and semantic segmentation purpose. It contains more than 200,000 HD images from videos and 20,000 HD images from snapshots under different styles of scenes including European style, modern city, highway, and green areas. The dataset provides groundtruth annotation for 13 class labels of semantic segmentation, depth estimation, and car ego-motion. Recently, MeteorNet \cite{MeteorNet} creates a Synthia 4D dataset derived from Synthia dataset. They generate 3D videos by back-projecting the depth image to the 3D space. The 6 sequences are selected under 9 weather conditions in different scenarios.

\noindent\textbf{Waymo Open Dataset (WOD)} \cite{sun2020scalability} is a recent large-scale self-driving dataset including two datasets including perception and motion dataset. The whole dataset contains 1,150  scenes where the training, validation and testing split consists of 798, 202, and 150 scenes respectively. The perception dataset is annotated with 1,950 lidar sequences while the motion dataset has 103,354 sequences. Each sequence is collected at sampling frequency of 10Hz and last 20s.


\section{Scene Flow Estimation} \label{sec:flow}

\begin{figure*}
  \centering\scriptsize
  \subfloat[Voxel-based representation.]{\includegraphics[width=0.2\textwidth]{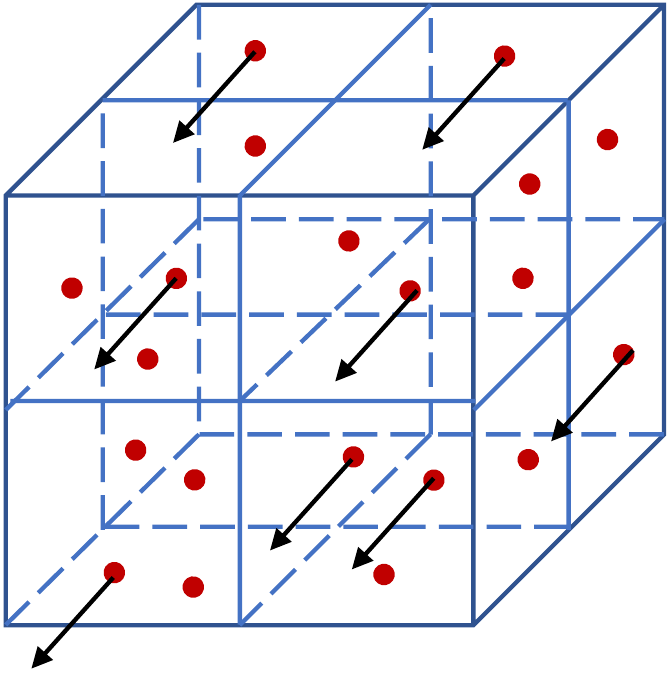}\label{fig:4_flow_a}}\hspace{13mm}
  \subfloat[Point-based representation.]{\includegraphics[width=0.2\textwidth]{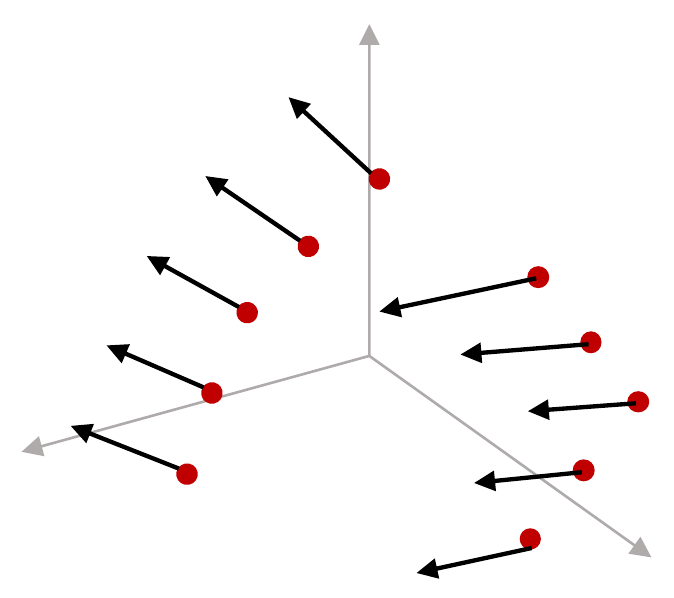}\label{fig:4_flow_b}}\hspace{15mm}
  \subfloat[Lattice-based representation.]{\includegraphics[width=0.2\textwidth]{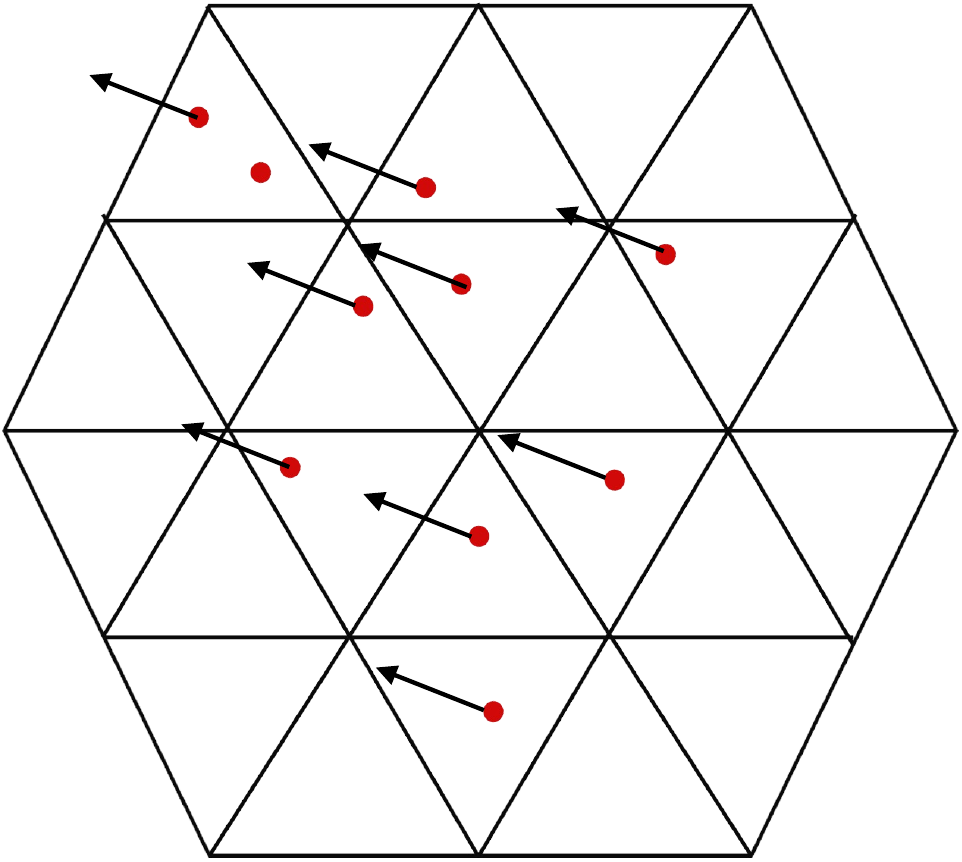}\label{fig:4_flow_c}}
  \caption{The illustration of different representations for scene flow estimation methods. Red points show the first point cloud frame while black arrows demonstrate related scene flow vectors.}
  \label{fig:scene_flow}
\end{figure*}

In dynamic SPL, scene flow estimation is one of the most crucial and fundamental tasks. It is playing a more and more important role in the applications of robotics manipulation, autonomous driving and so on. Flow actually describes the motion status of objects. Specifically in 4D point cloud, scene flow demonstrates  3D velocity of each 3D point in a scene. Assuming we have two consecutive point clouds in a point cloud sequence $S_t = \sum_{i=1}^{N_t} p_{i}^{t} $ and $S_{t+1} = \sum_{i=1}^{N_{t+1}} p_{i}^{t+1}$, scene flow $D_t = \sum_{i=1}^{N_t} D_{i}^{t} $ is defined as the translation motion vector between $S_t$ and $S_{t+1}$. For each point $p_{i}^{t}$ in $S_t$, the translated point is defined as $q_{i}^{t+1}$. $D_{i}^{t} = q_{i}^{t} - p_{i}^{t} $. It worth to note that $q_{i}^{t}$ and $p_{i}^{t+1}$ are not necessary to be the same location.

Based on the learning schema of networks, existing point cloud scene flow estimation methods can be divided into supervised and unsupervised methods. A list of scene flow estimation methods can be found in Table \ref{tab:sum:flow}.

\subsection{Supervised Methods}

These methods require groundtruth labels to train networks to obtain scene flow representation. According to the representation of data, these methods transfer SPL to different modalities (voxel, point, or lattice). 

\subsubsection{Voxel-based Methods}

These methods \cite{behl2018pointflownet,palafoxvoxflownet,jund2021scalable} voxelize SPL into volume representation and apply common 3D CNN to extract motion features. The scene flow is then computed among the centroids of these voxels. Figure \ref{fig:4_flow_a} demonstrates a typical voxel-based representation for scene flow estimation. The input point cloud was first divided into voxel representations and then processed by the regular 3D CNN networks such as VoxelNet \cite{VoxelNet}.
PointFlowNet \cite{behl2018pointflownet} jointly predicted the 3D object bounding boxes and their rigid motion results. They designed multiple branches of the decoder networks to jointly compute scene flow, ego-motion, and object detection results. Combining with  scene flow estimation and object detection results, rigid motions at both pixel level and object level could be obtained. Finally, multiple loss functions of various tasks were calculated together to optimize the network.
Similar to PointFlowNet \cite{behl2018pointflownet}, VoxFlowNet~\cite{palafoxvoxflownet} was another concurrent work that adopted voxel representation to estimate scene flow. However, VoxFlowNet took the farthest sampling strategy instead of  randomly sampling during the point selecting stage within a given voxel. Subsequently, they combined the idea from PointNet++ \cite{PointNet++} and FlowNet3D \cite{liu2019flownet3d}. First, local neighbor features were aggregated in each voxel to form the centroids feature. Then Set Conv layers were added to extract the features for centroids and Set Upconv layers were used to up-sampling voxels to the original scale to obtain the scene flow estimation results.

However, most of previous work were limited for the scale of point clouds. The computation cost would be extremely large with the scale increasing. Scalable \cite{jund2021scalable} was able to scale the size of  point cloud to $O$(100K) in a real-time system. Specifically, the authors built the network based on PointPillars \cite{PointPillars}. After the feature extraction, a dynamic voxelization method was applied followed by a U-Net autoencoder to process features. At last, Scalable recover the point-wise scene flow prediction  by shared MLP layers. Distinct from previous methods relying on the KNN search to find the neighbor regions which was super time-consuming, the pillar representation used in this paper saved a lot of computation cost to make the method scalable. 
Another major contribution of this paper was that the authors introduced a new scene flow estimation benchmark based on Waymo Open Dataset \cite{sun2020scalability}. The amount of real scene flow datasets with annotated groundtruth was very limited. The proposed new dataset provided a large possibility for people to conduct the scene flow estimation-related research. 

\begin{table*}[t]
  \centering
  \caption{The summary of scene flow estimation methods}
  \resizebox{\textwidth}{!}{
    \begin{tabular}{c|c|l||c|c}
    \hline 
    \multicolumn{3}{c||}{Methods} & Code  & Attribute \\
    \hline \hline
    \multirowcell{9}{Supervised} & \multirow{3}[2]{*}{Voxel-based} & PointFlowNet \cite{behl2018pointflownet} &  $\times$      & \multirowcell{4}{The voxel-based methods are convenient to apply the regular developed   \\ convolution networks but suffer from the quantization errors and lost  \\  information for correspondence matching.} \\
          &       & VoxFlowNet~\cite{palafoxvoxflownet} &  \checkmark    &  \\
          &       & Scalable~\cite{jund2021scalable} &  \checkmark      &  \\
          &        & PV-RAFT~\cite{wei2020pv}        &  \checkmark      &  \\
         
\cline{2-5}          & \multirow{4}[2]{*}{Point-based} & FlowNet3D ~\cite{liu2019flownet3d} & \checkmark        & \multirowcell{4}{The point-based methods are more nature to scene flow estimation task   \\   which is essentially point-wise prediction task. } \\
          &       & FlowNet3D++~\cite{FlowNet3D++}&    $\times$     &  \\
          &       & PointPWCNet~\cite{Wu2019PointPWCNetAC} &   \checkmark     &  \\
          &       & FESTA~\cite{wang2021festa} &    \checkmark    &  \\
          
\cline{2-5}          & Lattice-based & HPLFlowNet~\cite{gu2019hplflownet} &   \checkmark    & The lattice-based methods are more efficient and generalized well to various densities.   \\
    \hline
    \multirowcell{2}{Unsupervised} & \multirowcell{2}{Point-based} & Just Go \cite{mittal2020just}&   \checkmark    & \multirowcell{2}{The unsupervised methods release the labeling effort for large-scle point clouds \\ and  break the limit of insufficient scene flow datasets. }  \\
        &  & PointPWCNet \cite{Wu2019PointPWCNetAC}&  \checkmark     & \\ \hline
    \end{tabular}}
  \label{tab:sum:flow}%
\end{table*}%

\begin{table*}[ht]
  \centering
  \caption{Quantitative scene flow estimation results on FlyingThings3D \cite{Mayer2016ALD} and  KITTI \cite{KITTI:Scene:flow} datasets. End-Point-Error (EPE) computes the mean Euclidean distance between the ground-truth and the scene flow prediction. Acc Strict calculates the percentage of points with EPE $<$ 0.05m or relative error $<$ $5\%$; while Acc Relax calculates the percentage of points with EPE $<$ 0.1m or relative error $<$ $10\%$. $^{\star}$ indicates methods tested on datasets pre-processed by ~\cite{gu2019hplflownet}.}
  \resizebox{\textwidth}{!}{
    \begin{tabular}{c|c|l||ccc|ccc}
    \hline
    \multicolumn{3}{c||}{\multirowcell{2}{Methods}} & \multicolumn{3}{c|}{FlyingThings3D} & \multicolumn{3}{c}{KITTI} \\
\cline{4-9}    \multicolumn{3}{c||}{} & \multicolumn{1}{c}{EPE (m)} & Acc S. ($\%$) & Acc R. ($\%$) & \multicolumn{1}{c}{EPE (m)} & Acc S. ($\%$) & Acc R. ($\%$) \\ \hline
    \hline
    \multirowcell{8}{Supervised} & \multirow{2}[2]{*}{Voxel-based} & VoxFlowNet~\cite{palafoxvoxflownet} & 0.2971 & 11.36 & 33.46 & -     & -     & - \\
          &       & PV-RAFT$^{\star}$ ~\cite{wei2020pv} & 0.0461 & 81.68 & 95.74 & 0.056 & 82.26 & 93.72 \\
\cline{2-9}          & \multirowcell{5}{Point-based} & FlowNet3D ~\cite{liu2019flownet3d} & 0.1694 & 25.37 & 57.85 & 0.122 & 18.53 & 57.03 \\
          &       & PointPWCNet$^{\star}$ \cite{Wu2019PointPWCNetAC} & 0.0588 & 73.79 & 92.76 & 0.0694 & 72.81 & 88.84 \\
          &       & FlowNet3D++~\cite{FlowNet3D++} & 0.1369 & 30.33 & 63.43 & 0.253 & -     & - \\
          &       & MeteorNet~\cite{MeteorNet} & 0.2090 & -     & 52.12  & 0.2510 & -     & - \\
          &       & FESTA~\cite{wang2021festa} & 0.1113 & 43.12 & 74.42 & 0.0936 & 44.85 & 83.35 \\
\cline{2-9}          & Lattice-based & HPLFlowNet~\cite{gu2019hplflownet} & 0.1318 & 32.78 & 63.22 & 0.119 & 30.83 & 64.76 \\
    \hline
    \multirowcell{2}{Unsupervised} & Point-based & Just Go \cite{mittal2020just} & -     & -     & -     & 0.122 & 25.37 & 57.85 \\
\cline{2-9}          & Voxel-based & PointPWCNet \cite{Wu2019PointPWCNetAC} & 0.1246 & 30.68 & 65.52 & 0.2549 & 23.79 & 49.57 \\
    \hline
    \end{tabular}}
  \label{tab:results:flow}%
\end{table*}%

\subsubsection{Direct Point-based Methods}

However, voxel-based methods usually cost large computation resources and cannot guarantee the completeness of information. With these concerns, some methods \cite{liu2019flownet3d,FlowNet3D++,wang2021festa} take raw point clouds as input to directly estimate scene flow vectors.  We demonstrate the point-based representation in Figure \ref{fig:4_flow_b}, which computes scene flow vector for each point. These methods usually extract features in spatial domain and estimate the flow embedding in temporal domain. 
FlowNet3D~\cite{liu2019flownet3d} was a pioneer work by exploiting direct point-based representation for scene flow estimation. It builds the network upon the previous work PointNet++ \cite{PointNet++} to extract features for two consecutive point cloud frames in spatial domain. Specifically, the farthest point sampling  was adopted to sample the neighbor points and hierarchically aggregate the local features. The following flow embedding layer was used to concatenate the features of two point cloud frames including spatial locations and feature vectors. Finally, the flow was refined by the set upconv layers and output the upsampled flow point cloud. The whole network was end-to-end and achieved impressive performance on both the synthetic Flythings3D dataset and the KITTI Scene Flow 2015 dataset. Concurrent to  FlowNet3D, Shao et al. \cite{shao2018motion} proposed a method that jointly estimated scene flow, as well as  segmentation results and  motion trajectories. The network took two consecutive RGBD images as input and passed them to an encoder-decoder network. The flow estimation actually benefited from the semantic learning process, while FlowNet3D did not need any semantic supervision and directly predicted the flow based on the original point cloud.

Although FlowNet3D \cite{liu2019flownet3d} achieved the state-of-the-art performance, however, FlowNet3D used the naive $l2$ loss between the predicted scene flow and groundtruth annotation. So in the FlowNet3D++ paper, while still built the network upon FlowNet3D, the authors proposed two novel and powerful loss terms to constrain the flow estimation. One of them was the point-to-plane loss which was originally utilized for the point cloud registration problem. This loss can greatly improve the performance of the dynamic scene containing the deformable objects.  Another one was the cosine distance loss which solved the problem of the direction shifting between the predicted flow vectors and the ground truth. Further, the authors proposed a 3D dynamic reconstruction pipeline and integrated the scene flow estimation as a benchmark task. With the new evaluation measurement, the performance acquired a huge advancement compared to the FlowNet3D.

Almost all of the previous paper adopted PointNet++ \cite{PointNet++} as their feature extraction backbone. However, one major issue related to PointNet++ is the irregular sampling which leads to the randomness for feature extraction process. FESTA \cite{wang2021festa} used a spatial-temporal attention mechanism and achieved prominent benefits for scene flow estimation benchmarks. In the spatial domain, FESTA exploited a novel SA$^2$ layer to extract those points which were more stable and critical. The more representative points tended to help the network find better correspondence between the continuous frames. Likewise, in the temporal domain, FESTA introduced a TA$^2$ layer to tackle the various motion scale problem. A recurrent design was employed to first estimate an initial flow. Afterward, in the second iteration, FESTA shifted the attended region based on the initial flow which had more likelihood to find the good matches. The extensive experiments exhibited the significance of the proposed attention mechanism on scene flow estimation task. 

Pure point-based solution still concentrates on local correlations. The absence of global information leads to the error accumulation during previous coarse-to-fine strategies. Thus, the authors of \cite{wei2020pv} proposed a method named PV-RAFT applying point and voxel representations together to capture all-pairs correspondence. The K-NN pairs were adopted to model the local correspondence while pairs between volumes were utilized to involve global correlations. This improved the scene flow estimation performance especially for fast moving objects.

\subsubsection{Lattice-based methods}

Starting from PointNet \cite{PointNet} and PointNet++ \cite{PointNet++}, researchers always pre-process  point clouds and chunk them into small blocks before sending the data into the network. In this way, the global information is inevitably damaged and leads to inaccurate boundaries as well. 
Lattice-based methods splat point clouds into lattice space which could further leverage the Bilateral Convolutional Layers (BCL) \cite{jampani2016learning} to conduct scene flow feature learning. A typical lattice-based representation is shown in Figure \ref{fig:4_flow_c}.

Inspired from the Bilateral Convolutional Layers (BCL) \cite{jampani2016learning},  HPLFlowNet \cite{gu2019hplflownet} proposed a novel network which used the BCL and permutohedral lattice \cite{adams2010fast} to better estimate scene flow. The authors proposed DownBCL and UpBCL modified from the original BCL \cite{jampani2016learning} to extract the lattice features and refine scene flow from the coarse estimation respectively. Moreover, a CorrBCL was introduced to better fuse the information from two separate and consecutive point cloud frames. HPLFlowNet also presented a new density normalization schema which made the network much more efficient and was able to generalize to various point densities.

\subsection{Unsupervised Methods}

These methods eliminate the dependence on groudtruth labels by delicately designed unsupervised learning schema and also show promising performance which is even comparable with supervised methods. 

PointPWC-Net \cite{Wu2019PointPWCNetAC} is the first work that exploring unsupervised method to estimate scene flow in a coarse-to-fine manner  inspired by FlowNet \cite{ilg2017flownet} and PWC-Net \cite{sun2018pwc}. Specifically, to avoid the information loss in  previous single flow embedding layer such as in FlowNet3D method \cite{liu2019flownet3d}, the authors built a pyramid network for point cloud and hierarchically refine scene flow. At each pyramid level, they wrapped the first point cloud features with the up-sampled coarse flow from the last level, and computed the cost volume with the second point cloud features. Finally, the refined scene flow was acquired after the scene flow predictor. For supervised loss, they utilized the regular l2 loss for each layer between the groundtruth and the prediction. For the unsupervised loss, they introduced the Chamfer distance \cite{fan2017point}, smoothness constraint, and Laplacian regularization to predict the scene flow without any ground truth annotations. 

Just Go with the Flow \cite{mittal2020just} was another recent work that solely focused on using a unsupervised method and solving the lack of ground truth annotations in the real-world point cloud scene flow datasets. The authors built the network upon the FlowNet3D \cite{liu2019flownet3d} and introduced two loss functions to train the network. One was the nearest neighbor loss which was able to push the combination of the first point cloud and the forward flow towards the next point cloud. Another one was the cycle consistency loss which forced the combination of the next point cloud and the reverse flow to be close to the first point cloud. With these simple loss functions design, they could finetune the network on other large SPL data no matter whether they had the ground truth annotations and achieved the state-of-the-art performance.

\subsection{Discussion} 

The scene flow estimation results on both the synthetic dataset FlyingThings3D and real-world dataset KITTI are reported in Table \ref{tab:results:flow}. We have the following observations and discussions: 

\begin{itemize}
    \item Overall, the point-based and the lattice-based methods outperform the voxel-based methods by a large margin. This is because scene flow estimation is essentially a point-wise classification task. The dense representation such as point and lattice are naturally fit with the task, while the voxel representation might suffer from losing the fine-grained information. 
    
    \item Almost all types of methods demonstrate a well generalization ability from the synthetic domain to the real world domain. The models were trained on FlyingThings3D dataset and directly tested on KITTI dataset with promising performance. This reflects the potential of transfer learning and  few-shot learning prospects on more real applications. 
    
    \item The unsupervised methods still achieve comparable performance even without any supervision. This would largely solve the problem of not having enough annotated datasets in scene flow estimation task due to the large annotation cost. Any large-scale 3D sequential datasets could provide the training data for the unsupervised methods. Also, because of the problem definition, the second input frame could be naturally considered as the supervision. The unsupervised methods become a more and more popular research trend for the community.
    
\end{itemize}

\begin{table*}[ht]
  \centering
  \caption{The summary of multi-frame object detection methods.}
  \resizebox{\textwidth}{!}{
    \begin{tabular}{c|l||c|c}
    \hline
    \multicolumn{2}{c||}{Methods} & Code  & Attribute \\
    \hline \hline
    \multirow{4}[2]{*}{Convolution-based} & FaF \cite{FastandFurious}   &  \checkmark     & \multirowcell{4}{ The Convolution-based methods learn temporal information by  sliding window fusion \\ schema which is convolution operation. This is more convenient but tends to lose \\ details information or small objects. } \\
          & Second \cite{SECOND} &   \checkmark    &  \\
          & IntentNet \cite{casas2018intentnet} &   \checkmark    &  \\
          & What you see \cite{hu2020you} &    \checkmark    &  \\
    \hline
    Graph-based & Yin et al. \cite{yin2020lidar} & $\times$       &  The Graph-based methods benefit from spatial features extracted from graph networks.  \\
    \hline
    \multirow{3}[2]{*}{RNN-based} & YOLO4D \cite{el2018yolo4d} &   \checkmark    & \multirowcell{3}{ Compared to simple convolution operation, the RNN-based methods aggregate temporal \\ information better by exploring long-range temporal dependency. However, \\they usually cost more computation resources.} \\
          & McCrae et.al \cite{McCrae20203DOD} &  $\times$       &  \\
          & Huang et.al \cite{Huang2020AnLA} &    $\times$    &  \\
    \hline
    \end{tabular}}
  \label{tab:detection:mul}%
\end{table*}%

\section{Point Cloud Detection}
\label{sec:object_detection}

Object detection has been a significant computer vision task for a long time in both 2D and 3D domains which could bring tremendous applications such as self-driving, AR/VR, etc. The purpose is to recognize various objects and predict their precise bounding boxes in nature scenes. Previously, object detection in 2D images has made prominent achievements for both  efficiency or accuracy. Meanwhile, motivated by the success of 2D object detection, research about 3D object detection is driving more and more attentions in the community. However, most of them still concentrate on using single-frame data as input. Recently, some researchers started to apply methods by taking multiple frames, which is SPL data, as the input for networks. Temporal information is investigated to obtain boosted detection results on 4D (3D spatial and 1D temporal) sequential data such as point clouds. 
Compared to object detection methods with only using single point cloud as input, sequential 4D data is more appealing, since it provides much richer context information and wide-range coverage of temporal consistency. The real world scenes are often dynamic and hard to predict. Objects might be missing or occluded between continuous frames. Leveraging spatio-temporal information can significantly diminish false positives and false negatives during the object detection process. A list of multi-frame 3D object detection methods on SPL data is summarized in Table \ref{tab:detection:mul}.

\subsection{Convolution-based Methods} \label{sec:det_4d_conv}

\begin{figure}[ht]
\centering
\includegraphics[width=0.48\textwidth]{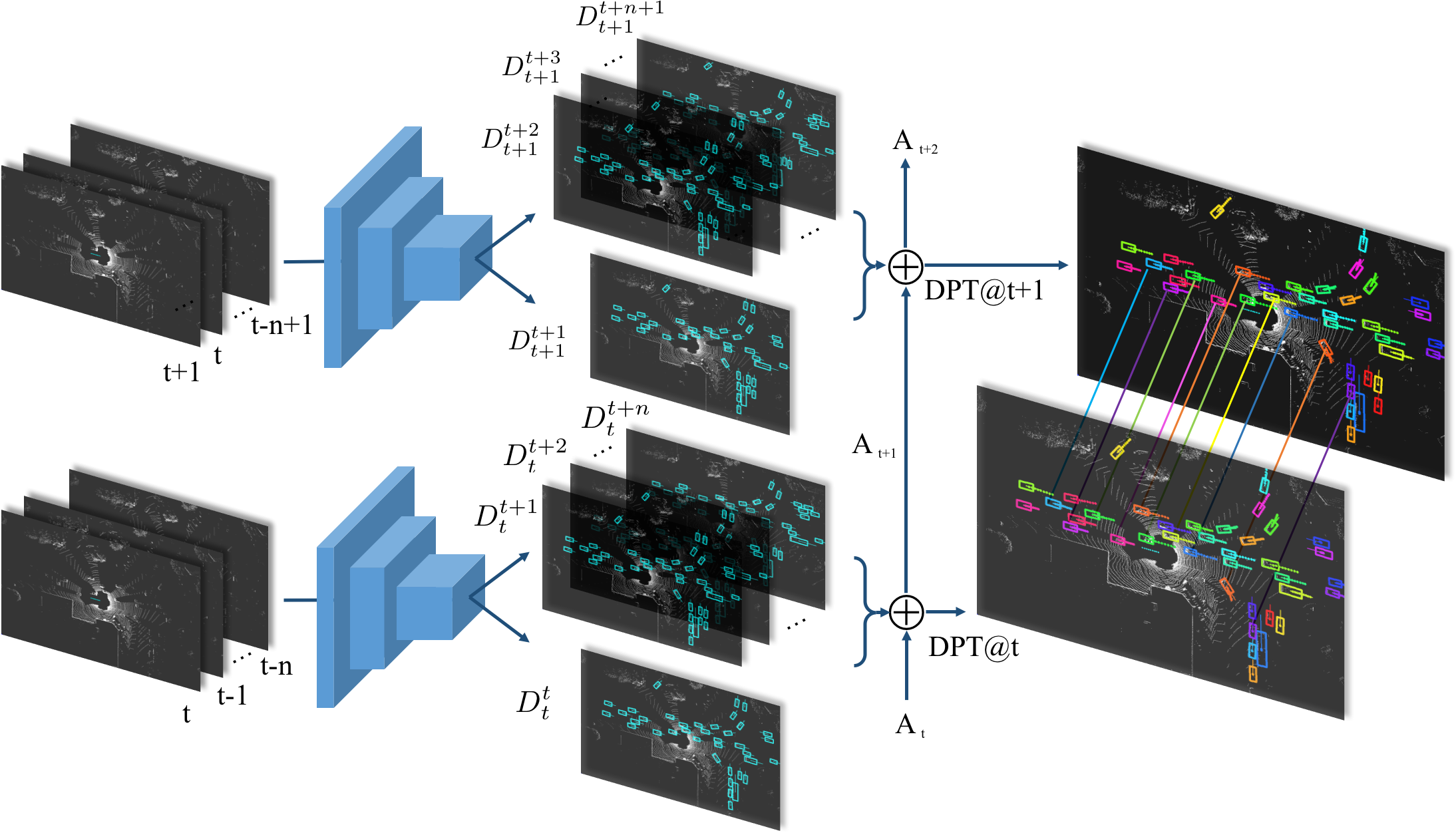}
\caption{The illustration of a convolution-based network for SPL object detection. The figure is from \cite{FastandFurious} with author’s permission.}
\label{fig:d:faf}
\end{figure}

\begin{table*}[ht]
\centering
\caption{Quantitative 3D object detection results on Waymo Open Dataset \cite{sun2020scalability} \emph{val} set (vehicles and pedestrians). 
}
\resizebox{\textwidth}{!}{
\begin{tabular}{cl||cccc|cccc}
\hline
\multicolumn{2}{c||}{\multirow{3}{*}{Method}}                                                                                                                             & \multicolumn{4}{c|}{Vehicles}                                                                                                                                                                        & \multicolumn{4}{c}{Pedestrians}                                                                                                                                                                      \\ \cline{3-10} 
\multicolumn{2}{c||}{}                                                                                                                                                    & \multicolumn{2}{c|}{3D AP}                                                                                  & \multicolumn{2}{c|}{BEV AP}                                                            & \multicolumn{2}{c|}{3D AP}                                                                                  & \multicolumn{2}{c}{BEV AP}                                                             \\ \cline{3-10} 
\multicolumn{2}{c||}{}                                                                                                                                                    & \multicolumn{1}{c|}{IoU=0.7}                         & \multicolumn{1}{c|}{IoU=0.8}                         & \multicolumn{1}{c|}{IoU=0.7}                         & IoU=0.8                         & \multicolumn{1}{c|}{IoU=0.5}                         & \multicolumn{1}{c|}{IoU=0.6}                         & \multicolumn{1}{c|}{IoU=0.5}                         & IoU=0.6                         \\ \hline \hline
\multicolumn{1}{c|}{\multirow{8}{*}{\begin{tabular}[c]{@{}c@{}}Single-frame \\ Methods\end{tabular}}} & StarNet \cite{REF:StarNet_2019}           & \multicolumn{1}{c|}{53.70}                           & \multicolumn{1}{c|}{-}                               & \multicolumn{1}{c|}{-}                               & -                               & \multicolumn{1}{c|}{66.80}                           & \multicolumn{1}{c|}{-}                               & \multicolumn{1}{c|}{-}                               & -                               \\
\multicolumn{1}{c|}{}                                                                                 & PointPillar \cite{PointPillars}   & \multicolumn{1}{c|}{60.25}                           & \multicolumn{1}{c|}{27.67}                           & \multicolumn{1}{c|}{78.14}                           & 63.79                           & \multicolumn{1}{c|}{60.11}                           & \multicolumn{1}{c|}{40.35}                           & \multicolumn{1}{c|}{65.42}                           & 51.71                           \\
\multicolumn{1}{c|}{}                                                                                 & MVF \cite{zhou2020end} & \multicolumn{1}{c|}{62.93}                           & \multicolumn{1}{c|}{-}                               & \multicolumn{1}{c|}{80.40}                           & -                               & \multicolumn{1}{c|}{65.33}                           & \multicolumn{1}{c|}{-}                               & \multicolumn{1}{c|}{74.38}                           & -                               \\
\multicolumn{1}{c|}{}                                                                                 & AFDET \cite{REF:AFDET_CVPRW2020}          & \multicolumn{1}{c|}{63.69}                           & \multicolumn{1}{c|}{-}                               & \multicolumn{1}{c|}{-}                               & -                               & \multicolumn{1}{c|}{-}                               & \multicolumn{1}{c|}{-}                               & \multicolumn{1}{c|}{-}                               & -                               \\
\multicolumn{1}{c|}{}                                                                                 & RCD \cite{REF:bewley2020range}             & \multicolumn{1}{c|}{68.95}                           & \multicolumn{1}{c|}{-}                               & \multicolumn{1}{c|}{82.09}                           & -                               & \multicolumn{1}{c|}{-}                               & \multicolumn{1}{c|}{-}                               & \multicolumn{1}{c|}{-}                               & -                               \\
\multicolumn{1}{c|}{}                                                                                 & PillarNet \cite{REF:PillarNet_ECCV2020}   & \multicolumn{1}{c|}{69.80}                           & \multicolumn{1}{c|}{-}                               & \multicolumn{1}{c|}{87.11}                           & -                               & \multicolumn{1}{c|}{72.51}                           & \multicolumn{1}{c|}{-}                               & \multicolumn{1}{c|}{78.53}                           & -                               \\
\multicolumn{1}{c|}{}                                                                                 & PV-RCNN \cite{shi2020pv}         & \multicolumn{1}{c|}{70.47}                           & \multicolumn{1}{c|}{39.16}                           & \multicolumn{1}{c|}{83.43}                           & 69.52                           & \multicolumn{1}{c|}{65.34}                           & \multicolumn{1}{c|}{45.12}                           & \multicolumn{1}{c|}{70.35}                           & 56.63                           \\
\multicolumn{1}{c|}{}                                                                                 & MVF++   \cite{Qi2021Offboard3O}                                            & \multicolumn{1}{c|}{74.64}                           & \multicolumn{1}{c|}{43.30}                           & \multicolumn{1}{c|}{87.59}                           & 75.30                           & \multicolumn{1}{c|}{78.01}                           & \multicolumn{1}{c|}{56.02}                           & \multicolumn{1}{c|}{83.31}                           & 68.04                           \\ \hline
\multicolumn{1}{c|}{\multirow{3}{*}{\begin{tabular}[c]{@{}c@{}}Multi-frames \\ Methods\end{tabular}}}  & Huang et al. \cite{Huang2020AnLA}              & \multicolumn{1}{c|}{63.60}                           & \multicolumn{1}{c|}{-}                               & \multicolumn{1}{c|}{-}                               & -                               & \multicolumn{1}{c|}{-}                               & \multicolumn{1}{c|}{-}                               & \multicolumn{1}{c|}{-}                               & -                               \\
\multicolumn{1}{c|}{}                                                                                 & MVF++  \cite{Qi2021Offboard3O}                                    & \multicolumn{1}{c|}{79.73}                           & \multicolumn{1}{c|}{49.43}                           & \multicolumn{1}{c|}{91.93}                           & 80.33                           & \multicolumn{1}{c|}{81.83}                           & \multicolumn{1}{c|}{60.56}                           & \multicolumn{1}{c|}{85.90}                           & 73.00                           \\
\multicolumn{1}{c|}{}                                                                                 & Qi et al. \cite{Qi2021Offboard3O}                                                & \multicolumn{1}{c|}{\textbf{84.50}} & \multicolumn{1}{c|}{\textbf{57.82}} & \multicolumn{1}{c|}{\textbf{93.30}} & \textbf{84.88} & \multicolumn{1}{c|}{\textbf{82.88}} & \multicolumn{1}{c|}{\textbf{63.69}} & \multicolumn{1}{c|}{\textbf{86.32}} & \textbf{75.60} \\ \hline
\end{tabular}}
\label{tab:d:waymo}
\end{table*}

\begin{table*}[ht]
\centering
\caption{{Quantitative 3D object detection results on nuScenes \cite{caesar2020nuscenes}  dataset. T.C. is the traffic cone. Moto. and Cons. represent the motorcycle and construction vehicle, respectively. }}
\resizebox{\textwidth}{!}{
\begin{tabular}{ccl||ccccccccccc}
\hline
\multicolumn{3}{c||}{Method}                                                                                                                                                                                                                  & Car                            & Pedestrian                     & Bus                            & Barrier                        & T.C.                           & Truck                          & Trailer                        & Moto.                          & Cons.                          & Bicycle                        & Mean                           \\ \hline \hline
\multicolumn{2}{c|}{\multirow{5}{*}{\begin{tabular}[c]{@{}c@{}}Single-frame \\ Methods\end{tabular}}}                                                          & VIPL\_ICT \cite{leaderboard}                          & 71.9                           & 57.0                           & 34.1                           & 38.0                           & 27.3                           & 20.6                           & 26.9                           & 20.4                           & 3.3                            & 0.0                            & 29.9                           \\
\multicolumn{2}{c|}{}                                                                                                                                          & MAIR \cite{simonelli2019disentangling}                & 47.8                           & 37.0                           & 18.8                           & 51.1                           & 48.7                           & 22.0                           & 17.6                           & 29.0                           & 7.4                            & \textbf{24.5} & 30.4                           \\
\multicolumn{2}{c|}{}                                                                                                                                          & PointPillars \cite{PointPillars}                      & 68.4                           & 59.7                           & 28.2                           & 38.9                           & 30.8                           & 23.0                           & 23.4                           & 27.4                           & 4.1                            & 1.1                            & 30.5                           \\
\multicolumn{2}{c|}{}                                                                                                                                          & SARPNET \cite{ye2020sarpnet}                          & 59.9                           & 69.4                           & 19.4                           & 38.3                           & 44.6                           & 18.7                           & 18.0                           & 29.8                           & 11.6                           & \{14.2\}                       & 32.4                           \\
\multicolumn{2}{c|}{}                                                                                                                                          & Tolist \cite{leaderboard}                             & 79.4                           & 71.2                           & 42.0                           & \textbf{51.2} & 47.8                           & \textbf{34.5} & 34.8                           & 36.8                           & 9.8                            & 12.3                           & 42.0                           \\ \hline
\multicolumn{1}{c|}{\multirow{9}{*}{\begin{tabular}[c]{@{}c@{}}Multi-frames\\ Methods\end{tabular}}} & \multicolumn{1}{c|}{\multirow{4}{*}{Convolution-based}} & FaF  \cite{FastandFurious}                                                                       & -                              & -                              & -                              & -                              & -                              & -                              & -                              & -                              & -                              & -                              & -                              \\
\multicolumn{1}{c|}{}                                                                                & \multicolumn{1}{c|}{}                                   & IntentNet  \cite{casas2018intentnet}                                                                  & -                              & -                              & -                              & -                              & -                              & -                              & -                              & -                              & -                              & -                              & -                              \\
\multicolumn{1}{c|}{}                                                                                & \multicolumn{1}{c|}{}                                   & Second  \cite{SECOND}                                                                      & -                              & -                              & -                              & -                              & -                              & -                              & -                              & -                              & -                              & -                              & -                              \\
\multicolumn{1}{c|}{}                                                                                & \multicolumn{1}{c|}{}                                   & What you see \cite{hu2020you}                         & 79.1                           & 65.0                           & 46.6                           & 34.7                           & 28.8                           & 30.4                           & 40.1                           & 18.2                           & 7.1                            & 0.1                            & 35.0                           \\\cline{2-14}
\multicolumn{1}{c|}{}                                                                                & \multicolumn{1}{c|}{\multirowcell{3}{RNN-based}}         & McCrae et al.  \cite{McCrae20203DOD}                                                              & 67.97         & 56.87                              & -                              & -                              & -                              & -                              & -                              & -                              & -                              & -                              \\
\multicolumn{1}{c|}{}                                                                                & \multicolumn{1}{c|}{}                                   & YOLO4D  \cite{el2018yolo4d}                                                                 & -                              & -                              & -                              & -                              & -                              & -                              & -                              & -                              & -                              & -                              & -                              \\
\multicolumn{1}{c|}{}                                                                                & \multicolumn{1}{c|}{}                                   & Huang et al.  \cite{Huang2020AnLA}                                                             & -                              & -                              & -                              & -                              & -                              & -                              & -                              & -                              & -                              & -                              & -                               \\\cline{2-14}
\multicolumn{1}{c|}{}                                                                                & \multicolumn{1}{c|}{Graph-based}                        & \textbf{Yin et al.} \cite{yin2020lidar} & \textbf{79.7} & \textbf{76.5} & \textbf{47.1} & 48.8                           & \textbf{58.8} & 33.6                           & \textbf{43.0} & \textbf{40.7} & \textbf{18.1} & 7.9                            & \textbf{45.4} \\ \cline{2-14} 
\multicolumn{1}{c|}{}                                                                                & \multicolumn{1}{c|}{Offboard detection}                     &   Qi et al. \cite{Qi2021Offboard3O}                                                                         & -                              & -                              & -                              & -                              & -                              & -                              & -                              & -                              & -                              & -                              & - \\ \hline
\end{tabular}}
\label{tab:d:nuscene}
\end{table*}

The convolution-based methods project SPL data into regular organization formats such as BEV (bird's eye view) map or voxel grid so that normal convolution operations could be leveraged to estimate object locations. A typical convolution-based network is shown in Figure \ref{fig:d:faf}.
FaF \cite{FastandFurious} jointly conducted 4D object detection, tracking, and motion forecasting together which took full advantage of multiple point cloud frames as input. These sub-tasks were shown to associate each other and boosted up the performance. Each point cloud frame was represented by voxel. Nevertheless, FaF did not perform 3D convolution on 3D voxel due to the large computation cost. Instead, it operated 2D convolutions on the $xy$ plane and directly treated the $z$ dimension  as feature information for 2D convolution. The same operation was applied for all of the frames and the coordinate system was normalized to be aligned across frames. The aggregated 4D tensor was sent to a single-stage object detector to accomplish the detection process. Meanwhile, to better utilize temporal information, FaF devised two schemes for temporal fusion. The early fusion directly concatenated tensors and used a 1D convolution to connect temporal features, while the late fusion hierarchically merged temporal features allowing the network to capture higher-level motion information. The object detection pipeline was the affinity of SSD \cite{yang20203dssd} mentioned above. Tracking and motion forecasting will be introduced in Sec. \ref{sec:object_tracking} and Sec. \ref{sec:forecasting-motion}. 

Yan et al. \cite{SECOND} introduced an improved sparse convolution on voxelized point cloud leading to faster computation. Likewise, an angle loss function was added to deal with the limited object orientation prediction problem. The authors aggregated temporal information by concatenating multiple point cloud frames and considering time stamps information as additional features for network's input. 
IntentNet \cite{casas2018intentnet} proposed a fully convolutional network to deal with object detection and intent prediction at a single pass. It represented 3D point cloud from bird’s eye view (BEV). The input data was modeled as 3D tensor and the height information was included as one of feature channels. Meanwhile, the temporal information from multiple Lidar sweeps was integrated into the height channel benefiting dynamic map and long trajectory predictions. 

Hu et al. \cite{hu2020you} argued that exploring free space for 2.5D data (RGBD or range image) is better than directly representing Lidar sweeps as 3D point clouds. The detection pipeline was built upon PointPillar \cite{PointPillars} architecture. The visibility map was derived through raycasting algorithm from voxelized input data, which can be further blended into the network gradient learning process. During training, the visibility volume was treated as an additional input to the network by two fusion methods, early fusion, and late fusion. The difference between these two fusion methods is located whether to compute input features separately using the backbone network. The aggregation of temporal information was considered to be an augmenting trick by taking the advantage of visibility prior. The authors of \cite{hu2020you} compensated motion by transferring SPL into a single scene and encoding timestamps as an additional input along with $xyz$ geometry, which can be proven to improve detection results by a large margin over PointPillar \cite{PointPillars} baseline model.   

\subsection{Graph-based Methods}

\begin{figure}[ht]
\centering
\includegraphics[width=0.48\textwidth]{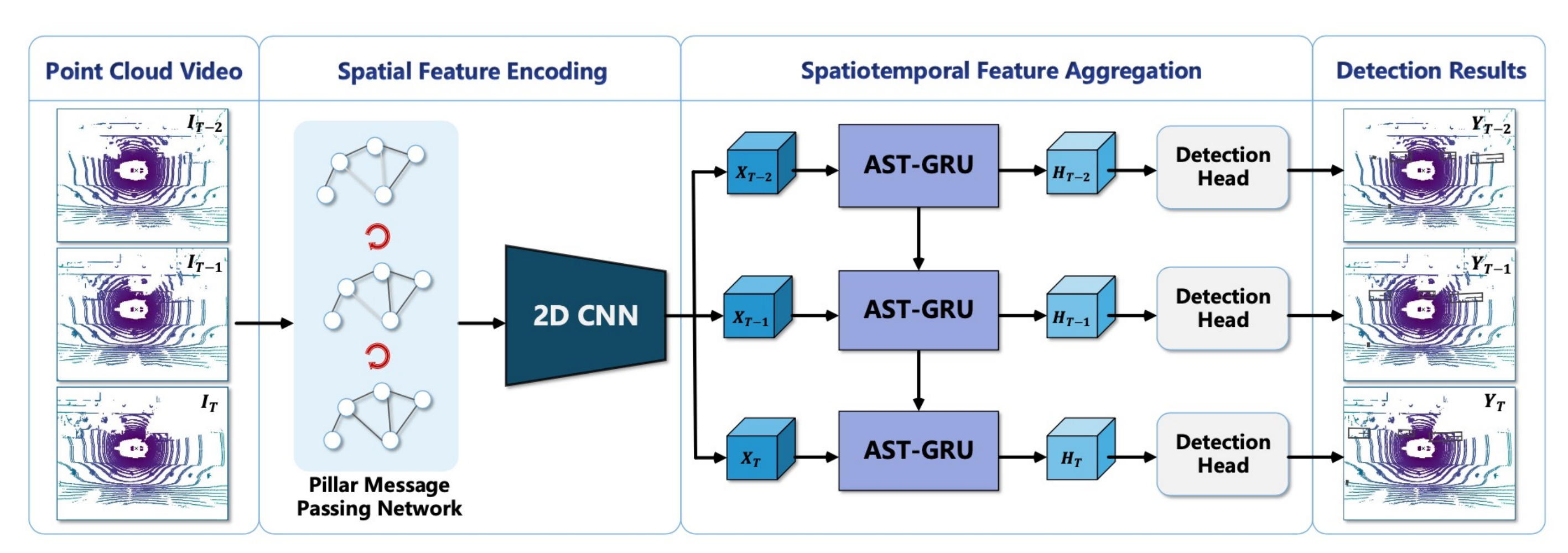}
\caption{The illustration of a graph-based method to conduct object detection. The figure is from \cite{yin2020lidar} with author’s permission.}
\label{fig:d:graph}
\end{figure}

The core idea of these methods is to explicitly capture point spatio-temporal strictures with graph networks modeling. Figure \ref{fig:d:graph} demonstrates a graph-based network to generate detection results. The network took SPL as input and all of the frames were aligned to the same coordinate system to eliminate ego-motion effects. After spatial features were extracted from point cloud frames, they were sent to Attentive Spatiotemporal Transformer Gated Recurrent Unit (AST-GRU) network to perform temporal information accumulation which can aid dynamic object detection results.  
Yin et al. \cite{yin2020lidar} explicitly proposed an object detection method from sequential point clouds and explored the superiority over single-frame 3D object detection which has limitations of sparse, occlusion and bias sampling, etc.  A delicate PMPNet was developed to manipulate the spatial relation from the encoded pillar grids graph in an iterative message-passing manner.


\subsection{RNN-based Methods}

\begin{figure}[ht]
\centering
\includegraphics[width=0.48\textwidth]{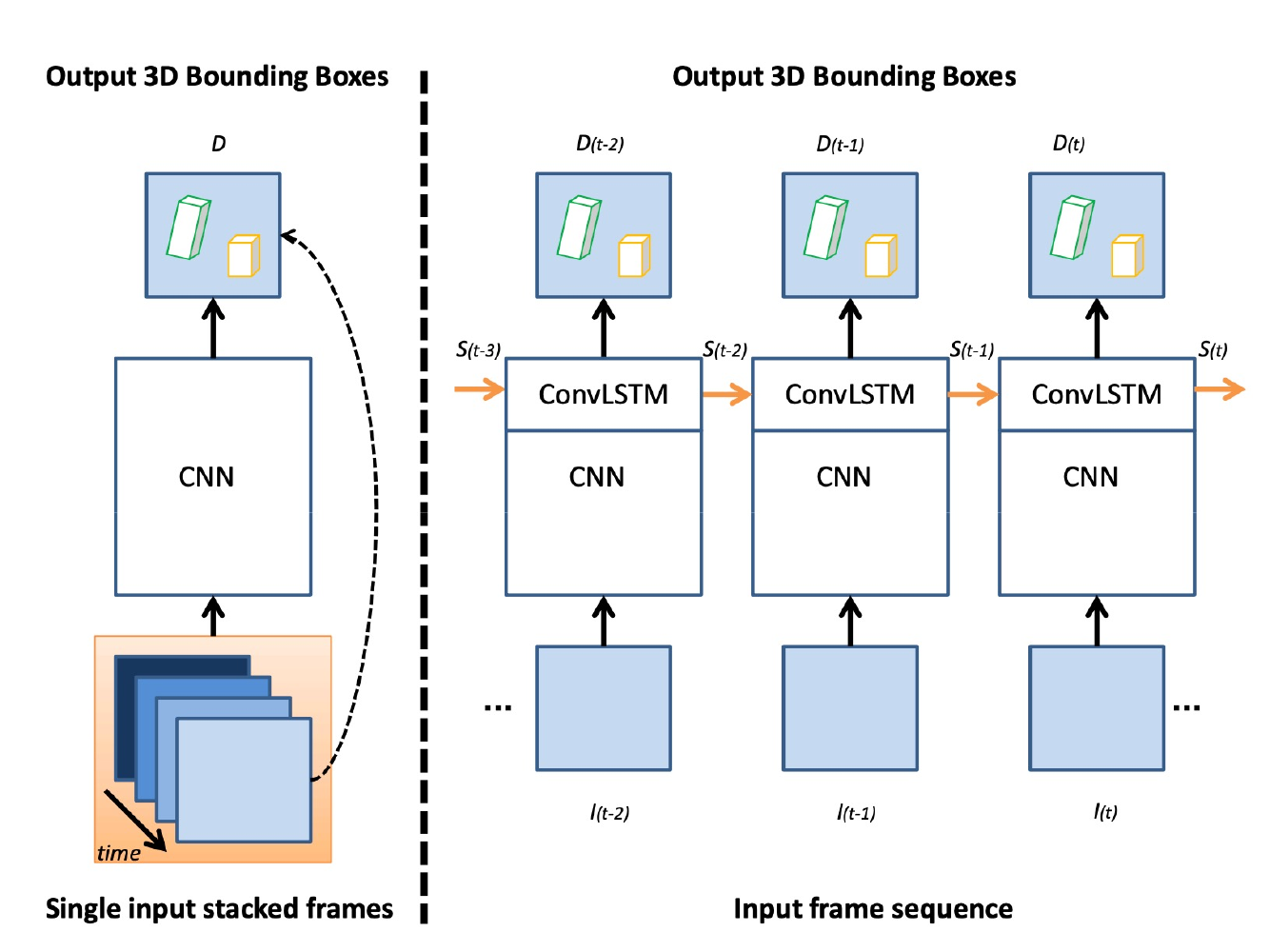}
\caption{The illustration of an RNN-based method for SPL objection detection. The figure is from \cite{el2018yolo4d} with author’s permission}
\label{fig:d:rnn}
\end{figure}

These methods \cite{Huang2020AnLA,el2018yolo4d,McCrae20203DOD} investigated recurrent neural networks to capture the temporal consistency of detection features and improved object localization accuracy. Figure \ref{fig:d:rnn} depicts a general idea of the RNN-based methods. The network  extracts spatial features by CNN for each point cloud frame. Then a recurrent network dubbed ConvLSTM is integrated to learn temporal features from previous state and current state, leading to generated features for the next layer.
The paper \cite{Huang2020AnLA} proposed by Huang et al. was the first one that modeled temporal relations among SPL with an RNN-based (LSTM) schema to boost up the performance of 3D/4D object detection results. The proposed network took SPL as input and generated backbone features for each point cloud frame by a 3D Sparse Conv U-Net. A novel 3D sparse LSTM was used to fuse backbone features across previous timestamp $t-1$ and current timestamp $t$. After embedding temporal information into hidden features, object proposals for each point were predicted by an object detection head network. Moreover, the authors built a knowledge graph among all of the point nodes to enhance spatial geometry information and suppress false positives. The final object detection results were refined by a traditional non-maximum suppression algorithm. 
 
Besides simple stacking LSTM layers which just concatenated SPL frames as 4D tensor and used CNN to comprehend temporal information, another way is to adopt powerful ConvLSTM.
YOLO4D \cite{el2018yolo4d} extended YOLO v2 \cite{REF:redmon2016you} network to 3D space and leveraged not only spatial but also temporal information from SPL. 
It could capture temporal information better and exhibited superiority during the multiple frames object detection process. 
McCrae et al. \cite{McCrae20203DOD} employed PointPillar \cite{PointPillars} as its baseline and developed a recurrent designed network that specifically takes three point cloud frames as input. Each point cloud frame was processed by a PointPillar model to extract features and followed a ConvLSTM to model temporal relation between the past and current time stamps. These designs were shown to be effective in pedestrian and vehicle classes.

\subsection{Offboard Detection}

Recently Qi et al. \cite{Qi2021Offboard3O}  explored an offboard application yielding groundtruth 3D labels by utilizing SPL detection results which have sufficient context information. Meanwhile, it reached the state-of-the-art 3D object detection performance on challenging Waymo Open Dataset. The multi-frames detection network dubbed MVF++ was built upon an advanced detection work MVF \cite{zhou2020end}. The authors followed the similar method of \cite{hu2020you} which aggregated temporal information by transforming other point cloud frames to the current one to get rid of ego-motion and encoded time offsets as an additional feature.

\begin{table*}[ht!]
  \centering
  \caption{The summarize of the Multiple Object Tracking methods. }
  \resizebox{\textwidth}{!}{
    \begin{tabular}{c|l||c|c}
    \hline
    \multicolumn{2}{c||}{Methods} & Code  & Attribute \\
    \hline \hline
    \multirowcell{5}{3D-based} & AB3DMOT \cite{weng20203d}  &   \checkmark     & \multirowcell{5}{The 3D based methods are more easy to implement and get rid of relying on other \\ data modalities. However, this usually are less sensitive to \\the extreme motion. } \\
     & FaF \cite{FastandFurious} &  \checkmark     &   \\
          & Chiu et al. \cite{chiu2020probabilistic}  &  \checkmark     &   \\
          
          & Giancola et al. \cite{3DSiamese} &   \checkmark      &  \\
          
          & PointTrackNet \cite{wang2020pointtracknet} &     \checkmark     &  \\
    \hline
    \multirowcell{5}{2D$\&$3D-based}
          & P2B  \cite{Qi2020P2BPN} &    \checkmark      & \multirowcell{5}{The joint 2D$\&$3D-based methods could be associative with the detection \\pipeline and are usually more accurate due to additional semantic signals from the 2D RGB \\ modality. However, the large computation cost is also inevitable. } \\
          & Complexer-YOLO  \cite{Complexer-YOLO-tracking} &    \checkmark      &  \\
          & DSM  \cite{frossard2018end} &    \checkmark      &  \\
          & GNN3DMOT \cite{weng2020gnn3dmot} &    \checkmark    &  \\
          
          & mmMOT \cite{zhang2019robust} &     \checkmark     &  \\
    \hline
    \end{tabular}}
  \label{tab:tracking}%
\end{table*}%

\begin{table*}[ht]
\centering
\caption{Quantitative 3D MOT Results of on KITTI Test Dataset. }
\label{tab:t:results}
\resizebox{0.8\textwidth}{!}{
\begin{tabular}{c|l||cccccc}
\hline
\multicolumn{2}{c||}{Method}                                                                                                          & MOTA$\uparrow$ & MOTP$\uparrow$   & MT$\uparrow$ & ML$\downarrow$ & ID\_sw$ \downarrow$ & FRAG$ \downarrow$ \\ \hline \hline
\multirow{5}{*}{\begin{tabular}[c]{@{}c@{}}3D-based\\ Methods\end{tabular}} & FaF \cite{FastandFurious}                                       & 80.9           & \textbf{85.3}             & 55.4         & 20.8           & -                   & -                 \\
& AB3DMOT \cite{weng20203d}                                   & 83.84          & 85.24            & 66.92        & 11.38          & \textbf{9}          & \textbf{224}      \\
                                                                                       & Chiu et al. \cite{chiu2020probabilistic}                           & -              & -                & -            & -              & -                   & -                 \\
                                                                                       & Giancola et al. \cite{3DSiamese}                             & -              & -                & -            & -              & -                   & -                 \\& PointTrackNet \cite{wang2020pointtracknet}  & 68.23        & 76.57          & 60.62     & 12.31        & \textbf{111}        & 725      \\
                                                                                      \hline
\multirow{4}{*}{\begin{tabular}[c]{@{}c@{}}2D$\&$3D-based \\ Methods\end{tabular}}    
                                                                                       & Complexer-YOLO\cite{Complexer-YOLO-tracking}    & 75.70       & 78.46          & 58.00      & 5.08         & 1186                & 2092              \\
                                                                                        & DSM\cite{frossard2018end}                 & 76.15       & 83.42         & 60.00      & 8.31        & 296                 & 868               \\
                                                                                       
                                                                                       & GNN3DMOT \cite{weng2020gnn3dmot}                                & 80.40          & 85.05            & 70.77        & 11.08          & 113                 & \textbf{265}               \\
                                                                                       & mmMOT \cite{zhang2019robust}                                   & \textbf{84.77}          & \textbf{85.21}            & 73.23        & \textbf{2.77}  & 284                 & 753              \\ \hline
\end{tabular}}
\end{table*}

\subsection{Discussion}
 
4D SPL object detection results on benchmarks of Waymo Open and nuScenes Datasets are summarized in Tables \ref{tab:d:waymo} and \ref{tab:d:nuscene}, respectively. Here are the observations and discussions: 

\begin{itemize}
    \item On both benchmarks of Waymo Open and nuScenes Datasets, the multi-frame methods demonstrate a clear superior performance compared to the single-frame methods. Although more information is involved, this does reflect the essence of additional temporal information. By using SPL data and devising spatio-temporal feature extracting techniques to conduct object detection, those false bounding box results are largely suppressed to ensure temporal consistency and thus improve overall detection accuracy.
    \item Compared to the RNN-based methods, the convolution-based and the graph-based methods accomplish better performance on nuScenes benchmark. As we also discussed in Sec. \ref{sec:network}, the RNN-based networks exploit more on temporal relations among long-range time series, while high-level semantic understanding tasks like detection prefer temporal consistency in both spatial and temporal domains.
    \item Almost all of the multi-frame detection methods are restricted to less than 10 frames. Thus long-range SPL object detection still remains as a challenging problem. 
\end{itemize}

\section{Point Cloud Tracking}
\label{sec:object_tracking}

4D multi-object tracking (MOT) is another essential application of SPL, which is also a vital component for autonomous driving task cooperated with 4D object detection prior. Being aware of object locations in each point cloud frame, 4D MOT takes the responsibility of associating them together in a whole sequence. The temporal consistency plays a crucial role to cope with the tracking problem in this process. Normally, 4D MOT system follows 2D MOT schema while the difference is the detection process happens in 3D space. In recent years researchers start to directly utilize 3D point cloud data to perform MOT even without any additional features such as RGB information. 

\subsection{3D-based Methods}

\begin{figure}[ht]
\centering
\includegraphics[width=0.48\textwidth]{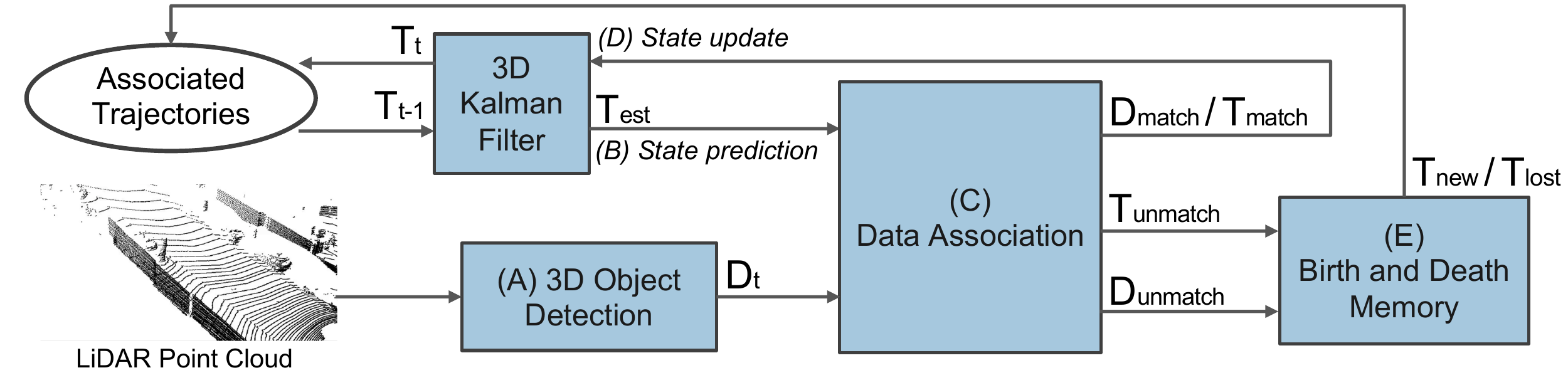}
\caption{A baseline for 3D-based MOT methods. The figure is from \cite{weng20203d} with author’s permission.}
\label{fig:t:baseline}
\end{figure}
Recent methods \cite{weng20203d,chiu2020probabilistic,FastandFurious,3DSiamese,wang2020pointtracknet} operate only raw SPL data for 3D MOT task. Usually these methods rely on object detectors to provide object locations and various filter-based algorithms to predict object trajectories, as shown in Figure \ref{fig:t:baseline}. 
AB3DMOT \cite{weng20203d} provided a compact baseline for multi-object tracking task while maintaining high-efficiency meeting the real-time estimation requirement. In this work, the authors derived detection results for the current Lidar frame through a pre-trained 3D object detector. The 3D Kalman Filter with constant velocity model predicted the state of object trajectory from previous frame. The predicted trajectory and detected objects were associated with Hungarian algorithm in current frame, which can further update trajectory state in 3D Kalman Filter. The authors also regularized the evaluation of 4D MOT system directly in 3D space instead of projecting into 2D plane as the previous work did. A new evaluation tool and three evaluation metrics were proposed to evaluate tracking performance on self-driving benchmarks in a more reasonable manner. Similar to the paper proposed by Weng et al. \cite{weng20203d}, Chiu et al. \cite{chiu2020probabilistic} dealt with the tracking problem using 3D Kalman Filter with a constant linear and angular velocity model. Besides the traditional approach, the authors exploited Mahalanobis distance for data association process and co-variance matrices for the state prediction process. 
 
However, previous filter-based methods were not sensitive to the extreme motion condition which may harm tracking performance. PointTrackNet \cite{wang2020pointtracknet} designed PointTrackNet to conduct object detection first from two continuous point cloud frames. The locations were further refined by an association model to merge detection results and ameliorate the impact of the false positive. The final tracklets can be provoked by linking matched objects. P2B \cite{Qi2020P2BPN} coped with the tracking problem with a point-wise schema and without using a traditional Kalman filter which has a relatively large computation cost. It proposed an end-to-end network and treated the tracking task as the detection task inspired by VoteNet \cite{VoteNet}. The sampled seeds and target centers embedded with local geometry information were first extracted from sequential point clouds. This strengthened the object representation instead of using single bounding box such as \cite{3DSiamese}. Then each target center was clustered with its neighbors to form the target proposal. Finally, object proposals were further verified over the whole sequence to ensure 3D appearance consistency and acquire tracking results.

Inspired by paper \cite{achlioptas2018learning}, to promote tracking performance with both richer feature representations and the regularization of the shape completion, Giancola et al. \cite{3DSiamese} proposed the first 4D MOT Siamese network structure. Specifically, first, the features extracted by an encoder network served as compact latent representations for Siamese tracker. Then the cosine similarity metric was used to match candidate shapes with model shapes. Finally, the decoder part of the shape-completion network was added to regularize Siamese tracker which could ensure the meaningful latent representation.

Different from the normal trajectory optimization solution, FaF \cite{FastandFurious} solved the tracking problem in an associative manner, incorporating with the object detection, motion forecasting and tracking tasks into a single pipeline. Firstly, as mentioned in Sec. \ref{sec:object_detection}, FaF adopted multi-frame object detection methods to derive object bounding boxes locations for the whole sequential frames. A motion forecasting algorithm was applied to predict object locations in further time stamps. In conjunction with past and current locations, tracklets could be obtained through average fusion.

\subsection{Joint 2D and 3D based methods} \label{sec:tracking-detction}

There are also some other methods \cite{frossard2018end,Complexer-YOLO-tracking,weng2020gnn3dmot} taking both 2D and 3D information together as input to further push the 3D MOT accuracy. In addition to SPL input, they involved another modality RGB image to the network as well. The features from different domains could complement each other and lean-to more representatives.
DSM \cite{frossard2018end} was an earlier work leveraging the deep structured model to create multiple neural networks together to solve the 4D MOT task. It predicted object proposals using a Detection Network from the input point cloud and RGB sequence. After formulating discrete trajectories, a liner optimization process was utilized to generate final tracking results.  

Unlike previous work such as \cite{weng20203d} extracting object features independently to perform the Hungarian data association, GNN3DMOT \cite{weng2020gnn3dmot} offered a novel multi-modality feature extractor to learn motion and appearance features from both 2D and 3D spaces. Furthermore, they firstly introduced a graph-based pipeline exploring the feature interaction among various objects to derive a more discriminate affinity matrix. Consequently, the data association process could benefit a lot from valuable object features which could also lead to a boosted tracking performance. 

To utilize high-level semantic features for 3D MOT task, the authors of \cite{Complexer-YOLO-tracking} generated semantic segmentation maps from input images. The semantic information was further back-projected to 3D space to obtain class-aware point clouds and provide extra semantic guidance to the tracking process. They predicted 3D bounding boxes from the voxelized semantic point cloud. The Scale-Rotation-Translation score (SRTs) was devised to reasonably evaluate performance and accelerate the speed to real-time.

\subsection{4D Single Object Tracking (SOT)}
Different from the widely used approach in MOT which links the tracking problem with the detection prior, 4D Single Object Tracking (SOT) aimed to estimate the object state in further frames based on the previous state. Pang et al. \cite{pang2021model} recently investigated 4D Single Object Tracking (SOT) and obtained tracklets through estimated object bounding boxes at various time stamps. The tracking process can be treated as a multi-frames registration method. Since there was no benchmark yet for the task, the authors contributed a new dataset LiDAR-SOT derived from Waymo Open dataset. 

\subsection{Discussion}

Table \ref{tab:t:results} summarizes 4D multi-object tracking (MOT) results on the KITTI benchmark. Several observations and discussions are listed below: 

\begin{itemize}
    \item Compared to pure 3D-based methods, joint 2D$\&$3D-based methods are more frequently used by the recent research community with a relatively higher performance, which shows the superiority of more modalities. 
    \item Most high-performance methods still require an additional 2D input to ensure tracking accuracy. This is a limitation with extra data. In the real self-driving scenario, usually, it costs much more to process multi-modalities at the same time. 
    \item For almost all of 3D MOT methods, tracking performance is based on detection performance. Only PointTrackNet \cite{wang2020pointtracknet} and P2B \cite{Qi2020P2BPN} belong to a full end-to-end pipeline breaking the limit of the off-shell detector. However, their performance is not satisfied which leaves a potential improvement for future research on this track. 
\end{itemize}

\section{Point Cloud Segmentation} \label{sec:object_segmentation}
Segmentation has always been another prevalent and crucial topic for high-level scene understanding including semantic segmentation, instance segmentation, and the combined version, panoptic segmentation. Distinct from detection and tracking, segmentation tasks demand a more fine-grained understanding of the surrounding scene. They require a pixel or point level classification for diverse scene object categories which could also provide a more holistic perception. Based upon previously developed 2D or single frame 3D segmentation methods, 4D segmentation over SPL recently gains amounts of popularity due to real applications in our dynamic world such as AR/VR, self-driving, etc. The path of handling the extra temporal dimension and keeping consistency in the 4D spatio-temporal space is paved by community. A list of SPL segmentation methods is summarized in Table \ref{tab:segmentation}.

In the following sections, we will cover 4D point cloud semantic and panoptic segmentation in Sec. \ref{sec:seg-sementic} and Sec. \ref{sec:seg-panoptic} respectively.

\subsection{4D Semantic Segmentation} \label{sec:seg-sementic}

The purpose of semantic segmentation is to apprehend semantic information from surrounding scenes and forecast the class label for each point in the point cloud. However, information provided by a single frame is usually insufficient. To get a relatively comprehensive perception of the real world, it is indispensable to explore approaches of fusing temporal information across multiple frames. 

\subsubsection{Simple Gathering}
\label{sec:seg:simple}

Some methods claim that the 4D semantic segmentation task can be simplified into the related 3D one.  Given SPL which have multiple frames, a network gathers point clouds into a single frame by transferring other frames' data into the coordinate system of the current frame. Then 3D semantic segmentation methods can be applied to solve the problem. 

\noindent\textbf{Projection-based} One large category of  3D semantic segmentation methods is the project-based methods. The input point clouds are primarily projected to the BEV (Bird’s Eye View) or the spherical space and then 2D segmentation pipelines can be easily applied to 2D projected data. Taking the advantage of advanced 2D CNN networks, the 3D segmentation process can be significantly sped
up. Zhang et al. \cite{zhang2018efficient} and PolarNet \cite{zhang2020polarnet} followed the BEV (Bird’s Eye View) projection track which format scene with a top-down snapshot. The network output segmentation results on the 2D spatial location including the semantic class prediction of the voxel along the $Z$-axis. Although these BEV project methods accomplished promising performance on segmentation benchmarks, scene information loss was inevitable. Spherical projection aimed to project point cloud data into the 360$\degree$ spherical space and then flatten it to the 2D image which can maintain maximum information. The resulted spherical projection image indicated structural information from the camera viewpoint. Studies \cite{wu2018squeezeseg,wu2019squeezesegv2,xu2020squeezesegv3,milioto2019rangenet++} followed the spherical projection track which treated the range image as the input data representation and predicted segmentation results with 2D CNN networks. In conjunction with some post-filtering technologies, 3D point cloud could be reconstructed from the range image. 

\noindent\textbf{Convolution-based} Researchers also represented 3D point cloud data with  regular grids so that 3D convolution operations could be applied to learn semantic features. Some studies \cite{huang2016point,segcloud,FCNN,Meng2018VVNetVV,rethage2018fully} transferred point cloud to voxel representation and adopted 3D convolutions over 3D volume data to estimate segmentation results for each occupancy grid. Although it was more straightforward to perform 3D semantic segmentation, 3D voxel convolution still suffered from the heavy computation cost and representation redundancy, leading to the inevitable accuracy and efficiency loss. Papers \cite{splatnet,rosu2019latticenet} splatted point cloud into the permutohedral lattice space to perform sparse convolutions. The lattice representation enables convolution operations to learn the semantic segmentation prediction while preserving maximum information at the same time. Octree \cite{schnabel2006octree} was another approach to formatting point cloud data. Octnet \cite{Riegler2016OctNetLD} was devised to conduct convolution operations on the octree structure for point cloud. PointConv \cite{wu2019pointconv} extended the convolution on the 2D image to the 3D domain with the dynamic filter which supported both the convolution and deconvolution.  KPConv \cite{kpconv} proposed Kernel Point deformable Convolution to cope with more flexible point cloud. 

\noindent\textbf{Point-based} Likewise, there is still another popular category directly processing 3D point clouds to estimate semantic segmentation results. Pioneered by  PointNet \cite{PointNet}, the authors proposed a shared-MLP based network and output point-wise labels for each point. Due to the lack of enough local geometry information,  PointNet++ \cite{PointNet++} attempted to add the grouping operation at multiple scales and resolutions to grab both local and global semantic features. Inspired by PointNet and PointNet++, a tremendous of point-based methods such as \cite{PointSIFT,know,PointWeb2019,zhang2019shellnet,randla} have been investigated to estimate semantic scene labels for point clouds. They exploited all kinds of different ways to aggregate representative features from local neighbors and promote segmentation performance. Some other methods such as \cite{all_you_need,Yang2019,LSANet,zhao2019pooling} introduced the attention mechanism to point-based networks to help extract more critical points and benefit segmentation results.  

\begin{table*}[htbp]
  \centering
  \caption{The summary of the sequential point cloud segmentation methods.}
  \resizebox{\textwidth}{!}{
    \begin{tabular}{c|c|l||c|c}
    \hline
    \multicolumn{3}{c||}{Methods} & Code  & Attribute \\
    \hline \hline
    \multirowcell{5}{Semantic \\ Segmentation} & \multirowcell{2}{Grid-based Convolution} & MinkNet \cite{MinkNet} &   \checkmark    & \multirowcell{2}{The grid-based convolution methods are more convenient to   implement due to the regular \\ gird representation of the point clouds, while inevitably suffer from the quantization error. } \\
          
          &       & SpSequenceNet \cite{Shi_2020_CVPR} &   \checkmark    &  \\
\cline{2-5}          & \multirowcell{2}{Point-based Convolution} & MeteorNet \cite{MeteorNet} &  \checkmark     & \multirowcell{2}{Point-based convolution preserve more information from the raw point clouds. }  \\

&       & PSTNet \cite{fan2021pstnet} &  \checkmark     &  \\
\cline{2-5}          & RNN-based Convolution & Duerr et al. \cite{duerr2020lidar}  &  $\times$     & Explicitly learn the temporal information but with higher computation cost.   \\
    \hline
    \multirowcell{2}{Panoptic \\Segmentation} & \multirow{2}[1]{*}{Point-based} & Ayg{\"u}n et al. \cite{aygun20214d}  &  \checkmark      & \multirowcell{2}{Jointly learning mutually boost each other and get a more \\holistic scene understanding.  }  \\
          &       & PanopticTrackNet \cite{hurtado2020mopt} &   \checkmark     &  \\ \hline
    \end{tabular}}
  \label{tab:segmentation}%
\end{table*}%

\begin{table*}%
\centering%
\caption{Quantitative semantic segmentation results on SemanticKITTI multiple scans dataset (IoU ($\%$)). The $\star$ shows moving classes.}%
\def\arraystretch{1.25}%

\scalebox{0.6}{
\begin{tabular}{ l || c | c c c c c c c c c c c c c c c c c c c c c c c c c}
  \hline
Methods & mIoU & \rot90{car } & \rot90{bicycle } & \rot90{motorcycle } & \rot90{truck } & \rot90{other-vehicle } & \rot90{person } & \rot90{bicyclist } & \rot90{motorcyclist } & \rot90{road } & \rot90{parking } & \rot90{sidewalk } & \rot90{other-ground } & \rot90{building } & \rot90{fence } & \rot90{vegetation } & \rot90{trunk } & \rot90{terrain } & \rot90{pole } & \rot90{traffic sign } & \rot90{{\text{car}$\star$}} & \rot90{{\text{bicyclist}$\star$}} & \rot90{{\text{person}$\star$} } & \rot90{{\text{motorcyclist}$\star$} } & \rot90{{\text{other-vehicle}$\star$} } & \rot90{{\text{truck}$\star$}} \\
  \hline \hline
  TangentConv \cite{tatarchenko2018tangent} & $34.1$ & $84.9$ & $2.0$ & $18.2$ & $21.1$ & $18.5$ & $1.6$ & $0.0$ & $0.0$ & $83.9$ & $38.3$ & $64.0$ & $15.3$ & $85.8$ & $49.1$ & $79.5$ & $43.2$ & $56.7$ & $36.4$ & $31.2$ & $40.3$ & $1.1$ & $6.4$ & $1.9$ & $\bm{30.1}$ & $\bm{42.2}$ \\
  DarkNet53Seg \cite{semantickitti} & $41.6$ & $84.1$ & $30.4$ & $32.9$ & $20.2$ & $20.7$ & $7.5$ & $0.0$ & $0.0$ & $91.6$ & $\bm{64.9}$ & $75.3$ & $\bm{27.5}$ & $85.2$ & $56.5$ & $78.4$ & $50.7$ & $64.8$ & $38.$1 & $53.3$ & $61.5$ & $14.1$ & $15.2$ & $0.2$ & $28.9$ & $37.8$ \\
  SpSequenceNet \cite{Shi_2020_CVPR} & $43.1$ & $88.5$ & $24.0$ & $26.2$ & $29.2$ & $22.7$ & $6.3$ & $0.0$ & $0.0$ & $90.1$ & $57.6$ & $73.9$ & $27.1$ & $\bm{91.2}$ & $\bm{66.8}$ & $\bm{84.0}$ & $\bm{66.0}$ & $\bm{65.7}$ & $50.8$ & $48.7$ & $53.2$ & $41.2$ & $26.2$ & $\bm{36.2}$ & $2.3$ & $0.1$ \\
  Duerr et al. \cite{duerr2020lidar} & $\bm{47.0}$ & $\bm{92.1}$ & $\bm{47.7}$ & $\bm{40.9}$ & $\bm{39.2}$ & $\bm{35.0}$ & $\bm{14.4}$ & $0.0$ & $0.0$ & $\bm{91.8}$ & $59.6$ & $\bm{75.8}$ & $23.2$ & $89.8$ & $63.8$ & $82.3$ & $62.5$ & $64.7$ & $\bm{52.6}$ & $\bm{60.4}$ & $\bm{68.2}$ & $\bm{42.8}$ & $\bm{40.4}$ & $12.9$ & $12.4$ & $2.1$ \\ \hline
\end{tabular}}
\addtolength{\tabcolsep}{1pt}  %
\label{tab:seg:kitti}%
\end{table*}%

\begin{table}[htbp]
\centering
  \caption{Quantitative semantic segmentation results on the Synthia 4D dataset.}
  \resizebox{0.48\textwidth}{!}{
\begin{tabular}{c|l||c|c|c}
\hline
\multicolumn{2}{c||}{Methods}                                        & Input & \#Params (M) & mIoU (\%)      \\ \hline \hline
\multirowcell{2}{Single \\ Frame} & 3D MinkNet14 \cite{MinkNet}         & voxel & 19.31        & 76.24          \\
                              & PointNet++ \cite{PointNet++}        & point & 0.88         & 79.35          \\ \hline
\multirowcell{4}{Multiple \\ Frames} & 4D MinkNet14 \cite{MinkNet}         & voxel & 23.72        & 77.46          \\
                              & MeteorNet \cite{MeteorNet}          & point & 1.78         & 81.8           \\
                              & PSTNet (l = 1) \cite{fan2021pstnet} & point & 1.42         & 80.79          \\
                              & PSTNet (l = 3) \cite{fan2021pstnet} & point & 1.67         & \textbf{82.24} \\ \hline
\end{tabular}}
\label{tab:seg:synthis}
\end{table}

\begin{table}[ht]
\centering
\caption{Quantitative 4D panoptic segmentation on SemanticKITTI validation set. MOT (Multiple Object Tracking) method by~\cite{weng2019baseline}; SFP (Scene Flow Propagation) Method by~\cite{mittal2020just}.}
\resizebox{0.48\textwidth}{!}{
\begin{tabular}{ll||ccc|cc}
\hline
\multicolumn{2}{c||}{Method}                                                               & LSTQ  & S$_\text{assoc}$ & S$_\text{cls}$ & IoU$^\text{St}$ & IoU$^\text{Th}$ \\ \hline \hline
\multicolumn{1}{l|}{\multirow{3}{*}{MOT}} & RangeNet++\cite{milioto2019rangenet++}        & 24.06 & 52.43            & 64.52          & 35.82           & 42.17           \\
\multicolumn{1}{l|}{}                     & KPConv~\cite{kpconv}                          & 25.86 & 55.86            & 66.90          & 47.66           & {54.13}         \\
\multicolumn{1}{l|}{}                     & Ayg{\"u}n et al. \cite{aygun20214d}               & 40.18 & 28.07            & 57.51          & 66.95           & 51.50           \\ \hline
\multicolumn{1}{l|}{\multirow{3}{*}{SFP}} & RangeNet++\cite{milioto2019rangenet++}        & 34.91 & 23.25            & 52.43          & 64.52           & 35.82           \\
\multicolumn{1}{l|}{}                     & KPConv~\cite{kpconv}                          & 38.53 & 26.58            & 55.86          & 66.90           & 47.66           \\
\multicolumn{1}{l|}{}                     & Ayg{\"u}n et al. \cite{aygun20214d} (1 scan) & 43.88 & 33.48            & 57.51          & 66.95           & 51.50           \\ \hline
\multicolumn{1}{l|}{}                     & Ayg{\"u}n et al. \cite{aygun20214d} (4 scans)     & 56.89 & 56.36            & 57.43          & 66.86           & 51.64      \\ \hline    
\end{tabular}}
\label{table:seg:panoptic}
\end{table}

\subsubsection{Temporal Convolution}

Simply gathering multiple frames into a single channel inevitably losses much spatial and temporal information  especially when there are large motions or deformations between frames. Instead of simply gathering, studies explored more advanced approaches \cite{MinkNet,Shi_2020_CVPR,MeteorNet,fan2021pstnet,duerr2020lidar} to learn the temporal information for the 4D semantic segmentation on sequential point clouds.

\noindent\textbf{Grid-based Convolution} These methods \cite{MinkNet,Shi_2020_CVPR} transferred point clouds to the regular data representation such as voxel occupancy and convolution operations could be applied along both spatial and temporal dimensions. Thus, the high-level context information could be fused across multiple frames and better inferring semantic perception in each frame. To achieve the point-wise semantic label prediction purpose, 4D MinkNet \cite{MinkNet} was the first method that applied the deep convolution network on high dimensional data such as SPL. It adopted the idea from Sparse Tensor \cite{graham2014spatially} and proposed the generalized sparse convolution to operate high dimensional data. The proposed convolution layer can be integrated with various deep networks and well generalized to different tasks. To deal with the computational problem when generalizing convolution to high dimensional spaces, the authors designed a novel kernel that is not hyper cubic and thus reduces the memory cost. The 4D segmentation network inherited the traditional 2D segmentation design U-Net \cite{ronneberger2015u} including sparse convolutions and sparse transpose convolutions. The skip connection was also adopted to link low-level and high-level layers. 

Although U-Net is a conventional method for semantic segmentation problem, its basic structure could still fail in some complex and dynamic scenarios. To better fuse global and local features, SpSequenceNet \cite{Shi_2020_CVPR} leveraged two novel models upon U-Net baseline to improve the segmentation performance, the Cross-frame Global Attention (CGA) and cross-frame local interpolation (CLI). 
The entire network structure took two consecutive frames as input and followed the U-net design in paper SSCN \cite{graham20183d} which contained 3D residual blocks in the encoder part. The Cross-frame Global Attention (CGA) model was utilized to import global attention information. It generated a mask from the previous frame which contained crucial semantic features such as appearance information. The mask could further guided the current frame feature extraction. Another model cross-frame local interpolation (CLI) was inspired by the scene flow embedding layer and fused both spatial and temporal feature information.

\noindent\textbf{Point-based Convolution} While grid-based methods are relatively consistent with 2D segmentation pipeline, they still suffer from quantization errors which lose information ineluctably. Compared with them, point-based convolution networks \cite{MeteorNet,fan2021pstnet} are usually more compact. They capture features from raw SPL data which reserve most object details information. MeteorNet \cite{MeteorNet} directly processed raw SPL data and performed spatio-temporal feature learning using a similar structure as PointNet++ \cite{PointNet++} which has been introduced in Sec. \ref{sec:net:point}. As for 4D semantic segmentation networks, MeteorNet built MeteorNet-Seg to conduct point-wise semantic label prediction process. The MeteorNet-Seg harnessed the Meteor-ind \cite{MeteorNet} module and the early-fusion strategy to construct the network. The Meteor-ind \cite{MeteorNet} module only contained neighbor points for each local patch due to point correspondence was not important for the segmentation task. The early-fusion strategy combined input point clouds early before the network to fuse temporal information. 

PSTNet \cite{fan2021pstnet} was another concurrent work designed for processing SPL with spatial-temporal convolution. The authors devised a Point tube structure to organize input data more efficiently and conduct proposed PST convolution. The point tube incorporated spatial and temporal kernels separately to capture spatio-temporal local structure information. To perform the point-level prediction task such as 4D semantic segmentation, the PST transposed convolution was developed to recover spatial and temporal scales which had been down-sampled by the PST convolution. Overall a hierarchical structure was built to process spatial and temporal features at different levels for 4D semantic segmentation task. Compared to grid-based methods such as \cite{MinkNet}, PSTNet was more compact yet effective while 4D MinkNet \cite{MinkNet} required a relatively large computation cost, especially with an increasing scale of data.

\begin{figure}[ht]
\centering
\includegraphics[width=0.48\textwidth]{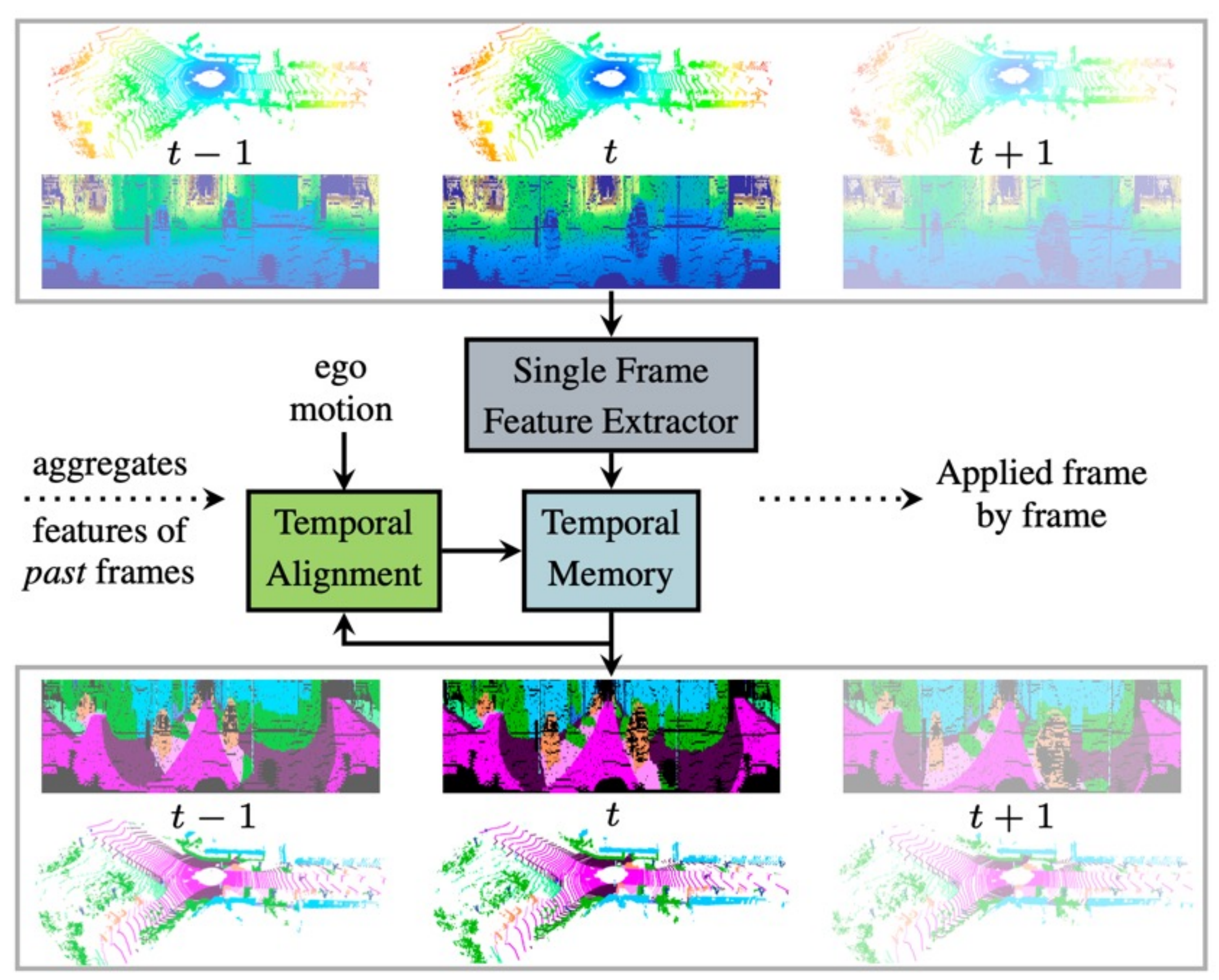}
\caption{The illustration of a RNN-based method for 4D semantic segmentation. The figure is from \cite{duerr2020lidar} with author’s permission.}
\label{fig:seg:rnn}
\end{figure}

\noindent\textbf{RNN-based Convolution} The RNN-based Convolution methods choose to aggregate temporal information recurrently as shown in Figure \ref{fig:seg:rnn}. Specifically, for each time stamp $t$, the network fused information from the previous frame at time $t-1$ and strengthened the segmentation of the current frame. The feature of the current frame would be continued to enhance future frames. 
Duerr et al. \cite{duerr2020lidar} projected each point cloud in a sequence to the image plane dubbed as range image mentioned in Sec. \ref{sec:seg:simple} and input to the network. For the entire sequence, the semantic feature would be perpetually reused instead of used just once in the previous paper such as SpSequenceNet \cite{Shi_2020_CVPR}. During temporal memory update process, the authors utilized two recurrent strategies to perform the feature fusion. One was adopting Residual Network which concatenates the past frame feature information with the current one and used MLP layers to conduct the spatial fusion. Another was ConvGRU dubbed as Gated Recurrent Unit which introduced gating mechanisms and replaced the MLP layer with the convolution layer. The latter one was a better choice which was able to achieve trade-off between efficiency and efficacy. 
  
\subsection{4D Panoptic Segmentation} \label{sec:seg-panoptic}

\begin{figure}[ht]
\centering
\includegraphics[width=0.48\textwidth]{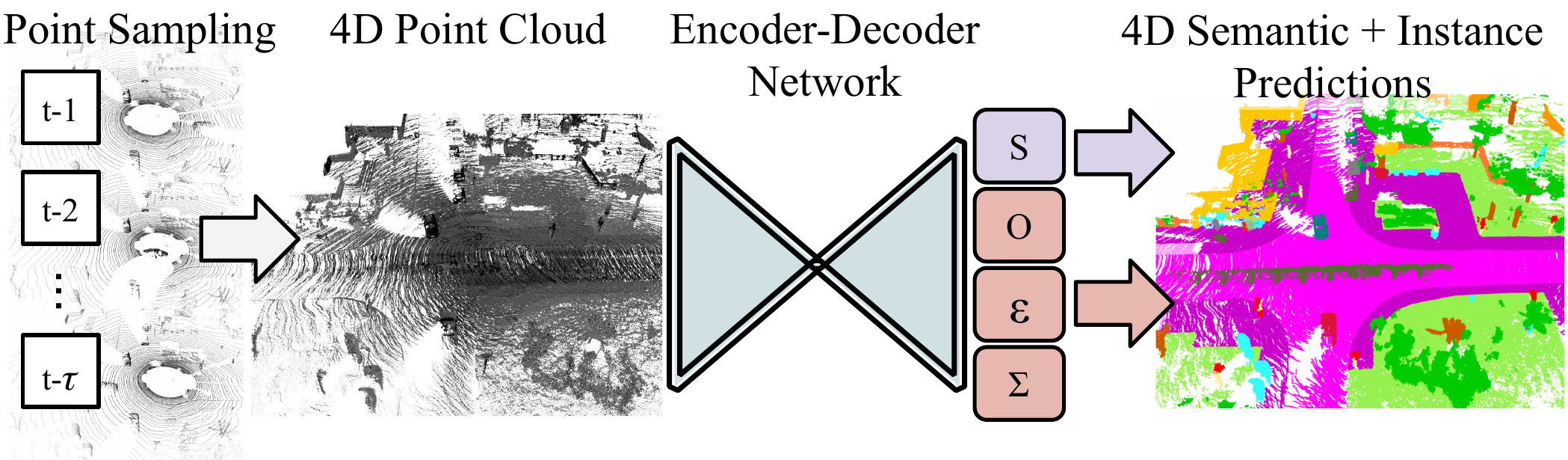}
\caption{The illustration of a typical 4D panoptic segmentation method. The figure is from \cite{aygun20214d} with author’s permission.}
\label{fig:seg:panoptic}
\end{figure}

Panoptic segmentation is a merged joint segmentation task including semantic segmentation and instance segmentation, which was first introduced in \cite{kirillov2019panoptic} in the image space and further extended from image to video by \cite{kim2020video}. Behley et al. \cite{behley2021benchmark} presented a large-scale Lidar benchmark for point cloud panoptic segmentation, in conjunction with baseline results for single-scan segmentation performance. 
Inspired from image to video upgrading in the 2D space and also the existing single-scan point cloud panoptic segmentation baseline, Ayg{\"u}n et al. \cite{aygun20214d} firstly proposed a 4D Panoptic Segmentation pipeline demonstrated in Figure \ref{fig:seg:panoptic}. The authors took a sequence of point clouds as input and inferred semantic classes for each point along with identifying the instance ID, completing both semantic and instance segmentation jointly for SPL. Different from the detection-based tracking paradigm Sec. \ref{sec:tracking-detction}, the authors first clustered points anchored on object center seeds and then assigned semantic information for each point. One major contribution in the paper was standardizing the evaluation protocol for the sequentially panoptic segmentation problem by devising a new point-centric evaluation method. Compared to existing metrics PQ~\cite{kirillov2019panoptic} and MOTA~\cite{voigtlaender2019mots} which had problems of over-estimating small segments and under-estimating frame association separately, the proposed LSTQ (LiDAR Segmentation and Tracking Quality) unified the evaluation in  space and time domains and measured point-to-instance association quality. 

To explore a holistic scene understanding problem,
PanopticTrackNet \cite{hurtado2020mopt} blended panoptic segmentation and multi-object tracking tasks. It proposed a novel architecture PanopticTrackNet with post-processing which unified semantic segmentation, instance segmentation, and multi-object tracking. The PanopticTrackNet was a multi-head end-to-end network containing a semantic segmentation head, instance segmentation head, and instance tracking head which simply concatenated frame vectors to merge temporal information. It took continuous RGB frames or point clouds as input and generated segmentation results. Then the MOPT fusion model was applied to predict the pixel-wise panoptic tracking output.

\subsection{Discussion} 

We summarize semantic segmentation results on the SemanticKITTI multiple scans benchmark and Synthia 4D dataset in Tables \ref{tab:seg:kitti} and \ref{tab:seg:synthis}, respectively. The results of 4D Panoptic Segmentation on SemanticKITTI \cite{semantickitti} dataset  are reported in Table \ref{table:seg:panoptic}. Based on these tables, we have the following observations and discussions: 

\begin{itemize}
    \item Additional temporal data improves the overall segmentation accuracy by a large margin compared to static point cloud methods as shown in Table \ref{tab:seg:kitti}, especially for those moving object classes. The motion information is well-captured by 4D semantic segmentation methods which further enhance the temporal consistency and remove false segmentation results. 
    
    \item From Table \ref{tab:seg:synthis}, point-based convolution outperforms grid-based convolution in terms of both efficacy and efficiency. Especially for efficiency, the number of parameters of point-based is much less than the grid-based methods, which avoids large computation cost of the quantization process.
    \item Overall segmentation performance is still limited on moving object classes which shows the large impact of motion information. 
    \item The panoptic segmentation methods significantly outperform other basic segmentation methods by exploring a holistic semantic scene understanding. The increase of scan numbers brings consistent performance gain. 

\end{itemize}

\section{Point Cloud Forecasting} \label{sec:forecasting}
Besides getting the perception of the surrounding world such as detection and segmentation, future forecasting is another critical component for a more holistic scene understanding. The reasonable and precise future prediction would largely decrease the uncertainty during  motion planning or self-driving process, especially in 3D space. Point cloud forecasting takes previous history information into the system and generates future object positions or entire scene point clouds, which would classify the task as motion forecasting or sequential forecasting. A list of point cloud forecasting methods is summarized in Table \ref{tab:forecasting}.    

In the following sections, the motion forecasting will be presented in Sec. \ref{sec:forecasting-motion} and the sequential forecasting will be summarized in Sec. \ref{sec:forecasting-sequentail}.

\subsection{Point Cloud Motion Forecasting} \label{sec:forecasting-motion}

\begin{figure}[ht]
\centering
\includegraphics[width=0.48\textwidth]{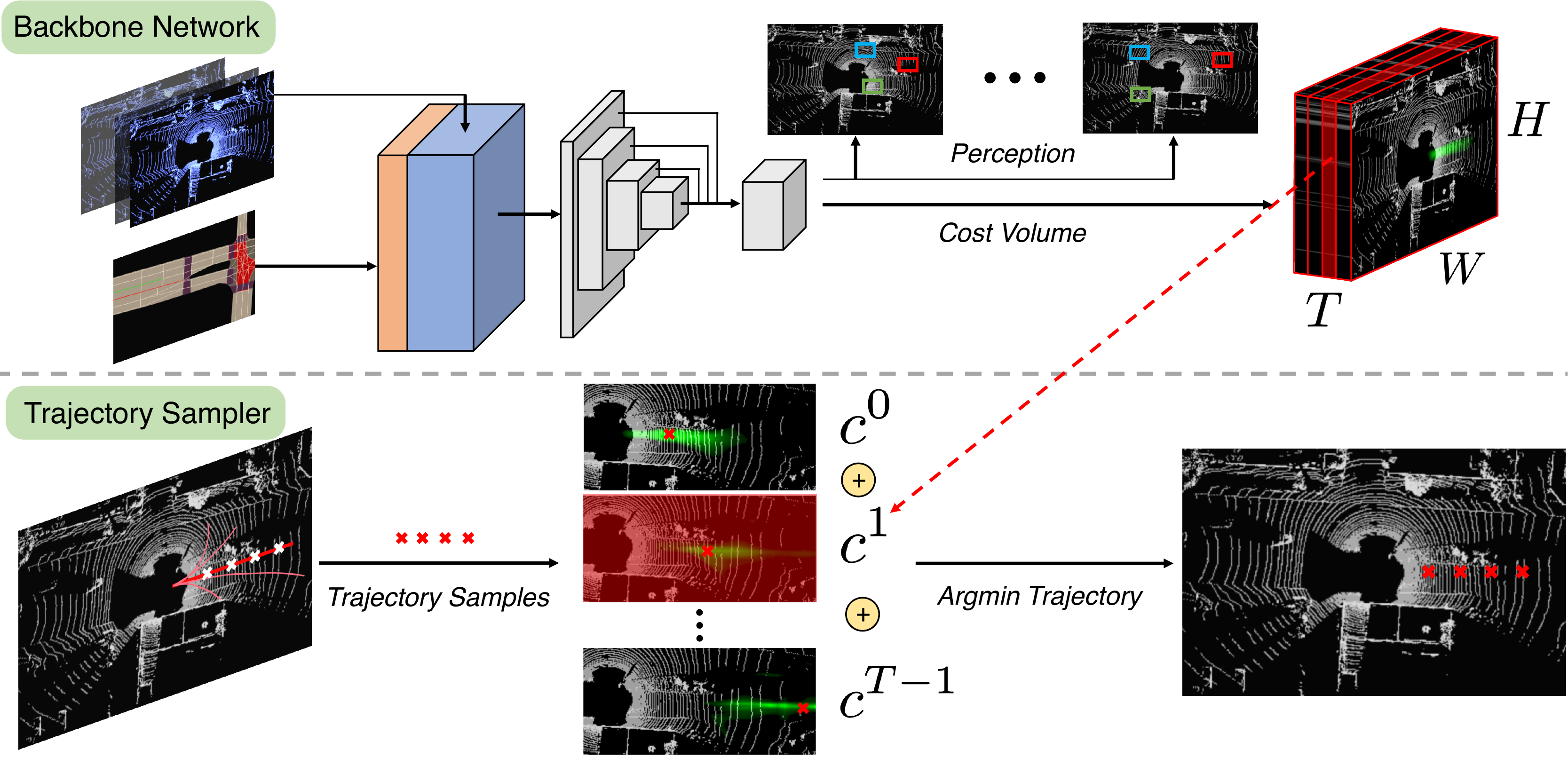}
\caption{The illustration of a voxel representation method for motion forecasting. The figure is from \cite{zeng2019end} with author’s permission.}
\label{fig:forecast:motion}
\end{figure}

Motion forecasting, also called motion prediction, aims to predict future object positions and trajectories by accumulating history spatial-temporal information. The conventional solution to this problem is usually associated with object detection and tracking, since knowing past object locations would provide strong prior knowledge to the future prediction. Usually, these methods are applied to image sequences or video signals by availing of powerful CNN networks. While high demands arise for predicting the future from raw sensor data, the community starts to explore motion forecasting from point clouds \cite{FastandFurious,casas2018intentnet,zeng2019end,casas2020spagnn,schreiber2019long,wu2020motionnet,meyer2020laserflow}.

\begin{table*}[htbp]
  \centering
  \caption{The summary of the sequential point cloud forecasting methods.}
    \begin{tabular}{c|c|l||c|c}
    \hline
    \multicolumn{3}{c||}{Methods} & Code  & Attribute \\
    \hline\hline
    \multirowcell{7}{Motion \\Forecasting} & \multirow{4}[2]{*}{BEV Representation} & FaF \cite{FastandFurious}   &   \checkmark    & \multirowcell{4}{The BEV Representation is more convenient to  implement due to the \\ regular projection which also makes the network more efficient} \\
          &       & IntentNet \cite{casas2018intentnet} &  \checkmark     &  \\
          &       & Spagnn \cite{casas2020spagnn} &   \checkmark    &  \\
          &       & NMP \cite{zeng2019end} &   $\times$    &  \\
\cline{2-5}          & \multirow{2}[2]{*}{OGM Representation} & Schreiber et al. \cite{schreiber2019long} &   $\times$    & \multirowcell{2}{The OGM Representation release the dependence on the\\ object detection results and improve the generalization ability.} \\
          &       & MotionNet \cite{wu2020motionnet} &   \checkmark    &  \\
\cline{2-5}          & Range View Representation & LaserFlow \cite{meyer2020laserflow} &   \checkmark    & Preserves more information from the raw point clouds.  \\
    \hline
    \multirowcell{5}{Sequential \\Forecasting} & \multirow{2}[1]{*}{Single-frame prediction} & Sun et al. \cite{sun2020novel}  &   $\times$    & \multirowcell{2}{These two methods are limited to the single \\ frame future prediction instead of sequential forecasting} \\
          &       & Deng et al. \cite{deng2020temporal} &   \checkmark       &  \\
\cline{2-5}          
         & \multirow{3}[2]{*}{Multi-frames prediction} & Weng et al. \cite{weng2020inverting} &   \checkmark      & \multirowcell{2}{The methods are adopting the range-view representation. Sun  et \\ al. \cite{sun2020novel} and Mersch et al. \cite{mersch2021self}  are limited to the deterministic prediction  \\ while S2net \cite{wengs2net} explore to extend the future uncertainty prediction. }  \\
          &       & Mersch et al. \cite{mersch2021self} &   \checkmark     &  \\
          &       & S2net \cite{wengs2net} &    \checkmark   &  \\ \hline
    \end{tabular}%
  \label{tab:forecasting}%
\end{table*}%

\begin{table*}[ht]
\centering
\caption{Quantitative detection and motion forecasting results on the NuScenes dataset. }
\label{tab:forecast:nuscene}
\begin{tabular}{c||c|ccc|cc}
\hline
\multirow{2}{*}{Method}             & Average Precision (\%) & \multicolumn{3}{c|}{$L_2$ Error (cm)}    & \multicolumn{2}{c}{Classification Accuracy ($\%$)} \\ 
                                    & 0.7 IoU                & 0.0 s       & 1.0 s       & 3.0 s        & MCA (Mean Category Accuracy)                   & OA (Overall Accuracy )                  \\ \hline\hline
Schreiber et al. \cite{schreiber2019long}                      & -                      & -           & -           & -            & 69.6                   & 92.8                   \\
MotionNet    \cite{wu2020motionnet}                       & -                      & -           & -           & -            & 70.3                   & 95.8                   \\ 
SpAGNN \cite{casas2020spagnn}       & -                      & \textbf{22} & 58          & 145          & -                      & -                      \\
LaserFlow \cite{meyer2020laserflow} & 56.1                   & 25          & \textbf{52} & \textbf{143} & -                      & -                     \\ \hline
\end{tabular}%
\end{table*}

\begin{table}[t]
    \centering
    \caption{Quantitative detection and motion forecasting results on the ATG4D dataset.}
    \begin{tabular}{c||c|ccc}
        \hline
        \multirow{2}{*}{Method} & Average Precision (\%) & \multicolumn{3}{c}{$L_2$ Error (cm)} \\ \cline{2-5}
        & 0.7 IoU & 0.0 s & 1.0 s & 3.0 s \\ \hline \hline
        FaF \cite{FastandFurious} & 64.1 & 30 & 54 & 180 \\
        IntentNet \cite{casas2018intentnet} & 73.9 & 26 & 45 & 146 \\
        NMP \cite{zeng2019end} & 80.5 & 23 & 36 & 114 \\
        SpAGNN \cite{casas2020spagnn} & 83.9 & 22 & 33 & \textbf{96} \\
        LaserFlow \cite{meyer2020laserflow} & \textbf{84.5} & \textbf{19} & \textbf{31} & 99 \\
        \hline
    \end{tabular}
    \label{tab:forecast:atg4d}
\end{table}

\subsubsection{BEV Representation}
Since point cloud data are usually sparse and irregular, one convenient and efficient way is adopting the Bird's Eye View (BEV) representation, which converts point clouds to 3D tensors \cite{FastandFurious,casas2018intentnet,zeng2019end,casas2020spagnn}. Besides the $XY$ location, height is treated as another feature to form one channel. In this way,  clear separations between target objects could still be preserved while largely reducing computation cost for high-dimensional data. Figure \ref{fig:forecast:motion} shows a typical BEV representation method for motion prediction. 

As introduced in Sec. \ref{sec:object_tracking} and Sec. \ref{sec:tracking-detction}, FaF~\cite{FastandFurious} was also the first one proposing a holistic network that jointly conducted object detection, tracking and motion forecasting from SPL input. Due to the association among multiple tasks, FaF had attained good fidelity for the motion prediction by adopting BEV representation. The IntentNet~\cite{casas2018intentnet} (introduced in Sec. \ref{sec:det_4d_conv}) extended FaF~\cite{FastandFurious} by predicting the intent which was defined as the combination of the target high-level behavior (e.g. moving directions) and motion trajectory. Besides SPL input, the authors took an extra rasterized map as network's input. The rasterized map consisted of the binary mask and poly lines which encoded static scene information including roads, traffic lights, traffic signs, etc. These signals provided a strong motion prior and contributed a lot to the intent prediction. The study  \cite{zeng2019end} further extended IntentNet~\cite{casas2018intentnet} to integrate motion planning into the end-to-end motion forecasting system. Instead of just predicting the moving angle as IntentNet~\cite{casas2018intentnet}, the purpose of motion planning was to generate one optimistic trajectory with minimum cost. Note that due to the novel joint design, multi-modality models were trained together in an end-to-end manner. The proposed motion planning was interpretable and generalized well to the uncertain situation. The \cite{casas2020spagnn} was also developed based on IntentNet~\cite{casas2018intentnet} by adding the interaction model at the end for motion predictions. It exploited a graph-based convolution neural network to model the relation between various actors and further decide the trajectory according to probabilistic inference. 

Nevertheless, these methods are all developed following the object detection-tracking-forecasting schema. The performance of the motion forecasting inevitably depend on the accuracy of bounding box positions derived from the first detection stage. If there are some unexpected objects failed to be detected or some unseen objects which are pretty normal in the real traffic situation, final forecasting results will be affected. 

\subsubsection{OGM Representation}

Occupancy grid map (OGM) was another popular representation for point cloud data. It partitioned the space into 2D grid cells with each cell indicating the occupancy and the point velocity of the space. The occupancy representation helped to predict the existence confidence of objects and thus did not need bounding boxes as the detection results.  Schreiber et al. \cite{schreiber2019long} was the one that adopted the occupancy grid map to forecast future motion for sequential raw sensor data. It converted point cloud frames to a sequence of dynamic occupancy grid maps and input them to a ConvLSTM encoder-decoder network to capture temporal dependencies. The ConvLSTM could predict future dynamic objects separating with the static scene. The authors added skip connections to the RNN network capturing multi-resolution features which could enhance the performance of the small object prediction. 

However, one major problem of the occupancy grid representation is hard to find the temporal correspondence between cells, which could further prevent better modeling behavior relations. Besides this, it also excludes object class information and sets the barrier for deeper analysis of the forecasted motion. Thus, MotionNet~\cite{wu2020motionnet} combined BEV and occupancy map representations and devised a novel representation named BEV map. It extended from the OGM and enriched the representation including the occupancy, motion, and object category information. After converting point cloud frames to a sequence of BEV maps, they were sent into MotionNet to obtain the scene perception and predict motion information. Specifically, MotionNet exploited a novel spatio-temporal pyramid network named STPN to extract hierarchical features and jointly modeled the space-time relations. Meanwhile, light block spatio-temporal convolution (STC) was developed to reduce computation cost of high dimension data and achieve real-time running.  

\begin{figure*}[ht]
\centering
\includegraphics[width=1\textwidth]{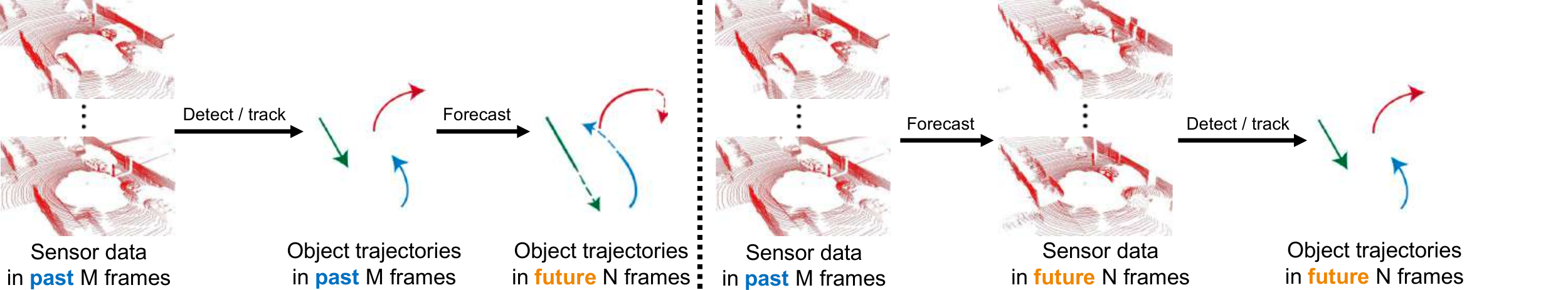}
\caption{Comparison of motion forecasting and sequential forecasting pipeline. The figure is from \cite{weng2020inverting} with author’s permission.}
\label{fig:forecast:sequential}
\end{figure*}

\subsubsection{Range View Representation}

Though two representations mentioned above could achieve promising performance for motion forecasting, they still suffer from quantization error and lose the information during the compression process. LaserFlow \cite{meyer2020laserflow} proposed to use the range view representation which provided more information than the BEV representation. As we also introduced in previous sections, the range map comes from spherical projection of point clouds.  LaserFlow \cite{meyer2020laserflow} treated multiple frames of range maps produced from point clouds as the input of the network. To aggregated multiple range maps, the multi-sweep fusion architecture was proposed to solve the coordinate system dis-alignment problem. In addition to extracting range map features, the authors exploited a transformer sub-network to unify the coordinate system and align all of the sweeps' features to the current one. The follow-by object detection and motion prediction network was applied to complete the motion forecasting by utilizing uncertainty curriculum learning. 

\subsection{Sequential Pointcloud Forecasting} \label{sec:forecasting-sequentail}

The SPF (Sequential Pointcloud Forecasting) task is defined to predict future \textit{M} point cloud frames given previous \textit{N} frames. Instead of forecasting future point cloud information on the object level,  SPF predicts the whole scene point clouds including foreground objects and background static scene. Also different from other generation tasks such as \cite{fan2017point,Lin:2018:LearningEP} mostly inferring the single point cloud frame \cite{sun2020novel}, SPF forecasts a sequence of future point cloud frames which requires longer temporal range information and more holistic scene understanding. Figure \ref{fig:forecast:sequential} demonstrates the difference between the motion forecasting and the sequential forecasting pipelines. 

Paper \cite{sun2020novel} aimed to resolve the point cloud compression and remove the redundancy part of spatial and temporal domains. It devised a ConvLSTM structure to predict future point cloud frames instead of using the 1D LSTM in \cite{weng2020inverting}. Deng et al. \cite{deng2020temporal} proposed a learning schema which adopted the scene flow embedding \cite{liu2019flownet3d} to model the temporal relation among four input point cloud frames. PointNet++ \cite{PointNet++} and Edge Conv \cite{Dynamic:Graph:CNN} were introduced to extract 3D spatial features. Combining spatial and temporal features, the network output the next future frame. However, the methods proposed by Sun et al. \cite{sun2020novel} and Deng et al. \cite{deng2020temporal} were limited to single future frame prediction setting while SPF requires a sequence of frames as inference results. 

Weng et al. \cite{weng2020inverting} firstly investigated the SPF (Sequential Pointcloud Forecasting) task and proposed a delicate method SPFNet which was able to predict the entire future 3D scene regardless of the human annotated ground-truth trajectories. The way it achieved this goal was through devising a novel forecast-then-detect schema to replace the conventional detect-then-forecast idea. In this way, all of the signals for the network training were future point cloud frames in a self-supervised manner. The proposed SPFNet employed the range map-based encoder and decoder structure to generate future point clouds. Meanwhile, a sequence of LSTMs was adopted to model the temporal relation among point cloud frames. The authors also exploited a new evaluation protocol that connected the detection and forecasting performance together to better assess the model. The SPFNet achieved the state-of-the-art performance on benchmark datasets compared to previous detect-then-forecast pipelines. 

Instead of leveraging the LSTM structure, Mersch et al. \cite{mersch2021self} proposed to utilize the 3D convolution to jointly learn spatial-temporal features of input point cloud sequences. 
It converted point clouds to range images which were then sent to an encoder-decoder network structure to extract features. Meanwhile, Skip Connections and Horizontal Circular Padding was introduced to capture detailed spatial-temporal information. Finally, the predicted future range images were converted back to sequential point clouds as output.

\subsection{Discussion} 

Tables \ref{tab:forecast:atg4d} and \ref{tab:forecast:nuscene} summarize results of motion forecasting on ATG4D and NuScenes datasets respectively. The observations and discussed can be found as follows: 

\begin{itemize}
    \item Though BEV representation is more frequently used, the methods adopting range view representation achieve better performance due to more complete information embedded. 
    \item Though existing motion forecasting methods have achieved remarkable performance on benchmarks, the errors  sharply increase when the time range is extended. This shows the limitation for handling longer-range SPL data. 
\end{itemize}

\section{Future Directions} \label{sec:future}

Sequential point clouds have been attracting great attention due to the need for a better and holistic scene understanding. Many methods have demonstrated the efficacy for processing high dimension data but with challenges and limitations. This section discusses some potential future research directions on the sequential point clouds.

\textbf{Unsupervised learning} Though there are a few unsupervised methods for the scene flow estimation, most existing research of sequential point clouds still rely on groundtruth labels as the supervision signal. The captured point clouds are easily scaled up to millions which bring an expensive cost to annotate groundtruth labels. Therefore, unsupervised learning schema will be a possible solution to release the requirement of annotation. Generally delicately designed unsupervised learning approaches  aim to learn meaningful semantic features from unlabeled data which could reduce the annotation cost. Obviously, large-scale unlabeled sequential point clouds can provide a great testbed for unsupervised learning methods. 

\textbf{Longer-range temporal dependency} Spatial feature learning has made great progress. The way how to capture and address temporal information is crucial for spatio-temporal learning. The existing research of sequential point clouds has attempted to model the temporal relation and leverage the long-range dependency to various applications such as tracking and forecasting. However, it is usually difficult to be accurate when the time range increases no matter for the input sequence or the output sequence. Another issue for a longer time range is the expensive computation cost due to a large amount of data. One possible solution is to exploit point cloud compression techniques such as utilizing flow information to fill the temporal gaps. Meanwhile, transformers have been approved to be quite good at modeling temporal attention and capturing long-range dependencies. Therefore, the combination of the two ideas could be an exciting future direction to model longer-range temporal dependencies. 

\textbf{Multitask Learning} Holistic perception of a scene is the foundation for applications of the sequential point clouds. Various tasks such as scene flow estimation, object detection and tracking, as well as segmentation, play an important role. For instance, scene flow estimation could provide the motion status of surrounding objects, while segmentation could deliver the object category information. However, by simply conducting these tasks separately, none of them could provide holistic guidance, while the results between tasks might even be inconsistent. Thus one possible solution is to jointly learn those essential features (e.g. semantic flow) across multitasks. For example, unified architectures could be designed to simultaneously learn scene flow and segmentation. The learned scene flow features and  semantic features could associatively boost each other while keeping the temporal consistency along the time sequence. Other multitask learning schemas are also worth devising especially for complex high dimensional data. 



\section{Conclusion} \label{sec:conclusion}

Deep learning for sequential point clouds has obtained a great success to better understand our dynamic world from spatio-temporal perspective and gained remarkable performance for various applications. We have provided a comprehensive survey of the recent deep learning methods for processing sequential point clouds as well as the downstream tasks. We believe this survey will
provide important guidance for researchers in computer vision and multimedia communities.

\small{
\bibliographystyle{plain}      
\bibliography{main}   
}

\end{document}